\documentclass{article}

\usepackage{arxiv}
\usepackage{algorithm, algorithmic}
\usepackage[utf8]{inputenc} 
\usepackage[T1]{fontenc}    
\usepackage{hyperref}       
\usepackage{url}            
\usepackage{booktabs}       
\usepackage{amsfonts}       
\usepackage{nicefrac}       
\usepackage{microtype}      
\usepackage{graphicx}
\usepackage{doi}

\usepackage{algorithm, algorithmic}

\usepackage{graphicx}
\usepackage{amsmath}
\usepackage{amssymb}
\usepackage{booktabs}
\usepackage{caption}
\usepackage{subcaption}
\usepackage{multirow}
\usepackage{balance}

\graphicspath{ {./images/} }

\title{Generating Adversarial Attacks in the Latent Space}

\author{ {\hspace{1mm}Nitish Shukla} \\
	Chennai Mathematical Institute\\
	Chennai, India \\
	\texttt{nitishs@cmi.ac.in} \\
	\And
	{\hspace{1mm}Sudipta Banerjee} \\
	IIIT- Hyderabad\\
	Hyderabad, India \\
	\texttt{sudipta.b@iiit.ac.in} \\
}

\renewcommand{\headeright}{}
\renewcommand{\undertitle}{}
\renewcommand{\shorttitle}{Generating Adversarial Attacks in the Latent Space}

\hypersetup{
pdftitle={Generating Adversarial Attacks in the Latent Space},
pdfsubject={q-bio.NC, q-bio.QM},
pdfauthor={Nitish Shukla, Sudipta Banerjee},
pdfkeywords={Adversatial Attacks, GAN},
}

\begin{document}
\maketitle
\begin{abstract}
Adversarial attacks in the input (pixel) space typically incorporate noise margins such as $L_1$ or $L_{\infty}$-norm to produce imperceptibly perturbed data that can confound deep learning networks. Such noise margins confine the magnitude of permissible noise. In this work, we propose injecting adversarial perturbations in the latent (feature) space using a generative adversarial network, removing the need for margin-based priors. Experiments on MNIST, CIFAR10, Fashion-MNIST, CIFAR100 and Stanford Dogs datasets support the effectiveness of the proposed method in generating adversarial attacks in the latent space while ensuring a high degree of visual realism with respect to pixel-based adversarial attack methods.
\end{abstract}

\section{Introduction}
\label{sec:intro}
Deep neural networks have been very successful in fields ranging from computer vision~\cite{10.1145/3065386, He2016DeepRL,Simonyan15}, reinforcement learning~\cite{DBLP:journals/corr/MnihKSGAWR13,mnih2015humanlevel} to natural language processing~\cite{10.5555/3295222.3295349,650093} and speech recognition ~\cite{Bahdanau2015NeuralMT}. Despite their success, robustness against noisy inputs is becoming a concern. Szegedy \textit{et. al.}~\cite{DBLP:journals/corr/SzegedyZSBEGF13} discovered that state-of-the-art deep learning models
suffer from vulnerability towards imperceptible input perturbations known as \textit{adversarial
attacks}.
Existing methods of generating adversarial attacks or adversarial samples such as Projected Gradient Descent (PGD)~\cite{madry2018towards} and Fast Gradient Sign Method (FGSM)~\cite{43405} follow the conventional procedure of adding minimal perturbations in the form of noise following a certain prior in the pixel (input) space. This prior is typically expressed as a $L_1$ or $L_{\infty}$ boundary or $\epsilon$-ball of radius
around the clean sample (per pixel) and serves as the attack budget or noise margin to be added to the pixels to render a successful adversarial attack. The attack margin can be interpreted as a pre-defined geometric prior, analogous to different forms of regularization used in optimization. Geometric prior-induced adversarial attacks in the input space are the most prevalent class of adversarial attacks in the literature. Note, the objective of adversarial attacks is to confound the deep learning network to make incorrect predictions. Although the adversarial perturbations are injected in the input space, the predictions are a result of operations in the high-dimensional feature space. This made us question, why not bypass the pixel space and launch the adversarial attack to directly affect the features in the latent space? Therefore, in this work, we explore a different approach of generating adversarial attacks with the following objectives. \textit{Firstly, to examine whether it is possible to exclude margin-based prior for generating adversarial attacks while maintaining high degrees of visual realism and secondly, to analyze whether it is feasible to directly induce adversarial perturbations in the latent space.} To conduct our investigation, we design an attack scheme independent of margin-based prior to induce adversarial perturbations in the \textit{latent space} that can generalize to both untargeted and targeted attacks.
 There are two advantages of injecting perturbations in the latent space. i) Perturbations in the latent space require minimal regulation on attack margin. The attack budget for the noise to be added need not conform to a specific bounded margin required in existing adversarial attacks. ii) Adversarial attacks in the latent space can provide intuitive cues to interpret and explain adversarial attacks. Our contributions in this work are as follows.

\begin{itemize}
    \item We use a generative adversarial network (GAN)~\cite{10.5555/2969033.2969125} to generate adversarial samples in the latent space that is generalizable to both targeted (specific target class) and untargeted (an arbitrary class) attack scenarios on several well-known image datasets, MNIST, CIFAR10, Fashion-MNIST, CIFAR100 and Stanford Dogs datasets. 
    \item We interpret the adversarial attacks in the latent space from a geometric perspective using convex hulls. We demonstrate how perturbations in the latent space push the original data towards the closest face of the convex hull of target class in the case of targeted attacks, and to the nearest class in the case of untargeted attacks. 
    \item We visualize the features extracted from the discriminator and observe that both original images and adversarial images form well-separated distributions supporting the feasibility of producing adversarial attacks in the latent space.
    \item We use class activation maps~\cite{Selvaraju_2019} to depict that original and perturbed images are highly complementary in the feature space while being similar in the pixel space.

\end{itemize}

The rest of the paper is organized as follows. \autoref{sec:rel} outlines the existing work. \autoref{sec:method} describes the rationale behind proposed method and the methodology for achieving untargeted and targeted attacks. \autoref{sec:exp} describes the experimental settings. In \autoref{sec:find}, we present our findings and analyze the results. \autoref{sec:sum} concludes the paper.

\section{Related Work}
\label{sec:rel}

\textbf{Gradient based adversarial attacks: } Some of the most successful adversarial attacks have been gradient based \textit{i.e.}, these methods leverage the gradients with respect to the input to find the adversarial samples. Attack schemes like fast gradient sign method (FGSM) and projected gradient descent (PGD) are classical examples of gradient based attacks. PGD attack can be viewed as an iterative version of FGSM where an attack is performed for a fixed number of iterations or until misclassification is achieved. These attacks are called \textit{white-box} attacks as they involve computation of the gradient of the model's weights.
While other methods focus on adding perturbations in the input holistically, attacks like jacobian-based saliency map attack (JSMA)~\cite{7467366} and one-pixel attack~\cite{Su_2019} restrict the perturbation within a small region in the image. Essentially, these methods select and change the pixels that are most-likely to produce the largest increase (largest gradient) in the loss. This process is repeated for a set number of iterations or until the data is misclassified.
Other methods like DeepFool attack~\cite{7780651} minimizes the norm of the adversarial perturbation by solving an optimization problem. To generate adversarial attacks, the iterative method pushes the image towards the closest hyperplane after linearizing the class boundaries around the current image to produce a convex polyhedron. The additive perturbation updates the image iteratively until it converges to a successful adversarial attack. AdvGAN~\cite{ADVGAN} uses a GAN to learn and approximate the distribution of original samples and then generate adversarial samples efficiently for each instance. It uses a user-specified bound in the form of soft hinge loss to restrict the magnitude of perturbation. AWTM~\cite{AWTGAN} uses a GAN and an autoencoder as a mapper to achieve attack without a target model and relies on two weight parameters to regulate the strength of the attack and the quality of the generated sample.
All of the above methods perform incremental bounded perturbation in the pixel space to deliberately change the classification decision. However, we propose an alternate scheme that can perturb the data in the latent space that results in adversarial attack while maintaining the perceptual quality of the data.

\section{Methodology}
\label{sec:method}

\subsection{Rationale}
\label{sec:ration}

In \cite{yousefzadeh2020deep}, the author analyzes convex hulls of the features produced by a deep classification network $f$, to demonstrate generalizability of deep learning-based methods. Inspired from the above work, we explore the effects of adversarial attacks in the latent (feature) space than in the input (pixel) space. Specifically, we develop a framework that is capable of perturbing the features of the original data so that it changes the class label without altering the visual semantics of the data. Therefore, we use a generative adversarial network (GAN) that is capable of synthesizing realistic images using an encoder-decoder architecture (generator) and incorporates a classifier (discriminator) to distinguish between original and perturbed samples. \textbf{In our case, the generator simulates an autoencoder~\cite{8616075}, to ensure that the original and perturbed images are almost identical. The discriminator serves the dual purpose of distinguishing between `real' (original) and `generated' (adversarially perturbed) samples, and their respective class labels. Therefore, the discriminator guides the generator to synthesize samples that are successfully perturbed, i.e., they must belong to different classes.} The proposed method is different from Latent Poison attack~\cite{LatentPoison}, that explores the vulnerability of variational autoencoders from a security perspective and requires $\epsilon$-bound on the class decision. An extension of the Latent Poison work in achieving only untargeted attacks by generating out-of-distribution samples is done in~\cite{SPL}. \textbf{However, we propose to perturb the features in the latent space for generating realistic images while attacking the class predictions (without noise margin) in an end-to-end fashion to accomplish both untargeted and targeted attacks.}

\subsection{Proposed Method}
\label{sec:prop}

We use a GAN~\cite{10.5555/2969033.2969125} to produce images that fools a classifier. The GAN consists of an encoder-decoder architecture as the \textit{generator} that generates adversarial samples, and a \textit{discriminator} that distinguishes between original and adversarial samples. We denote the generator and discriminator network by $\mathcal{G}$ and $\mathcal{D}$ respectively. Let $\textbf{x}=\{x^{(i)}\}_{i=1}^m$ denote a batch of $m$ samples, $\{y^{(i)}\}_{i=1}^m$ denote the corresponding labels, and $\tilde{\textbf{x}}=\mathcal{G}(\textbf{x})$, denote a set of generated images. During training, weights of both networks are optimized to minimize their respective loss functions. The discriminator's loss $\mathcal{L}_{\mathcal{D}}$ is formulated as

\begin{equation}
    \label{eq:d_loss}
    \begin{aligned}
    \mathcal{L}_{\mathcal{D}}= \lambda_1 \cdot \frac{1}{m}\sum\limits_{i=1}^m \mathcal{L}_{ce}(\mathcal{D}(\textbf{x}^{(i)}), \textbf{y}^{(i)}) \\+
    \lambda_2 \cdot \frac{1}{m}\sum\limits_{i=1}^m \mathcal{L}_{ce}(\mathcal{D}( \mathcal{G}(\textbf{x}^{(i)}) ), \tau)
    \end{aligned}
\end{equation}
and the generator's loss $\mathcal{L}_{\mathcal{G}}$ is formulated as
\begin{equation}
    \label{eq:g_loss}
    \mathcal{L}_{\mathcal{G}}= 
    \gamma_1 \cdot \frac{1}{m}\sum\limits_{i=1}^m \mathcal{L}_{ce}(\mathcal{D}( \mathcal{G}(\textbf{x}^{(i)}) ), \tau) + \gamma_2 \cdot \mathcal{L}_1 (\textbf{x},\tilde{\textbf{x}})
\end{equation}

In order to inject adversarial perturbations in the latent space, we strategically employ the generator and discriminator loss functions to supervise the adversarial sample generation. Note that classification occurs as a result of highly non-linear operations in the latent space. Therefore, we introduce loss terms that deliberately affect the classification decision by implicitly perturbing the features that are responsible for classification. We design the loss functions such that they regulate realistic data generation and accomplish adversarial attacks synchronously. 

\autoref{eq:d_loss} and \autoref{eq:g_loss} achieve the desired objective as follows. In the equations, $\tau$ denotes the target class in the case of targeted attack, $\mathcal{L}_1$ is per-pixel loss and $\mathcal{L}_{ce}$ is the cross-entropy loss. In case of untargeted attack, we replace the target class $\tau$ by $c^*-y^{(i)}$ where $y^{(i)}$ is the label of sample $x^{(i)}$ and $c^*=\max_{i}(c_i)$. This essentially forces the discriminator to misclassify a sample of class $c$ as $c^*-c$. When the number of classes in the dataset is even and class labels are 0-indexed, labeling class $c$ as $c^*-c$ mislabels all the training samples. 

Given a target class $\tau$, the first term in $\mathcal{L}_{\mathcal{D}}$ penalizes  misclassification of a sample $x^{(i)}$ to its class $y^{(i)}$ whereas the second term encourages $\mathcal{D}$ to classify the generated sample $\tilde{x}^{(i)}=\mathcal{G}(x^{(i)})$ as $\tau$. Similarly, the first term in $\mathcal{L}_{\mathcal{G}}$ encourages the generated images to be classified as $\tau$ whereas the second term forces the generated images to look visually similar to real images. Successful training results in a generator network that produces realistic images that appear to be sampled from original dataset, and a discriminator network that correctly classifies real images but classifies the generated images as belonging to $\tau$. Algorithm \autoref{alg:gan} outlines the procedure followed in the proposed method.

\noindent\textbf{Baseline: } We design a baseline for comparison as existing methods perturb in the pixel space. It consists of an autoencoder that is jointly trained with a fully connected layer to classify the embedding produced by the encoder. Let $h$ and $g$ be the encoder and decoder respectively, and $f'$ be the fully connected layer. The model minimizes the following loss.
\begin{equation}
    \label{eq:baseline_loss}
    \begin{aligned}
    \mathcal{L}_{\mathcal{B}}= \mu_1 \cdot \frac{1}{m}\sum\limits_{i=1}^m \mathcal{L}_{1}(g(h(\textbf{x}^{(i)})),\textbf{x}^{(i)}) \\+\mu_2 \cdot \frac{1}{m}\sum\limits_{i=1}^m \mathcal{L}_{ce}(f'(h(\textbf{x}^{(i)})),\textbf{y}^{(i)})
    \end{aligned}
\end{equation}
The first term in $\mathcal{L}_{\mathcal{B}}$ forces the model to produce images that look similar to the input whereas the second term supervises correct classification of the embedding produced by $h$.
Given a sample $x$ belonging to class $c$, we first construct the convex hull using embedding produced by $h$ of all the training samples from class $c$. We then iteratively move towards the direction of the nearest face in the hull for a fixed number of iterations or until the class label, \textit{i.e.}, the classification decision changes. 
\section{Experiments}
\label{sec:exp}

\textbf{Setup and implementation details: }
We use a ResNet-$18$~\cite{He2016DeepRL} as the discriminator network whose final layer is modified to output $C$ scores corresponding to $C$ class labels for each dataset used in this work (for example, $C$=10 for MNIST, Fashion-MNIST and CIFAR10 datasets, $C$=100 for CIFAR100, and $C$=120 for Stanford Dogs dataset). We follow the experimental settings in DCGAN~\cite{https://doi.org/10.48550/arxiv.1511.06434} for the generator. Specifically, we use the generator and discriminator of the DCGAN architecture as decoder and encoder respectively. We use Adam optimization~\cite{DBLP:journals/corr/KingmaB14} with a mini-batch size of $100$ and an initial learning rate of $0.002$ that is maintained using exponential-LR~\cite{Li2020An} scheduler to train the generator and the discriminator. Additionally,the weight parameters in discriminator's loss $\mathcal{L}_{\mathcal{D}}$ are set to $\lambda_1=0.5$ and $\lambda_2=0.5$, whereas in generator's loss $\mathcal{L}_{\mathcal{G}}$, they are set to $\gamma_1=0.1$ and $\gamma_2=0.9$. The discriminator weights are updated multiple times for each update of generator's weights to incorporate the slow training of $\mathcal{D}$. 
 We also initialize the weights of both the generator and the discriminator using He uniform initialization~\cite{article}. To select the model with best adversarial attack success rate, we do $k$-fold cross validation~\cite{DBLP:journals/corr/abs-1811-12808} with the value of $k$ set to $5$.
 
\algsetup{indent=0.5em}
\begin{algorithm}[t]
\caption{Proposed training algorithm. 
}
\label{alg:gan}
\begin{algorithmic}[1]
\REQUIRE  batch size $m$, target class $\tau$, max class $c^*$, discriminator $\mathcal{D}$, generator $\mathcal{G}$, and hyperparameters: $\lambda_1$, $\lambda_2$, $\gamma_1$, $\gamma_2$.
\medskip
\WHILE{training has not converged}
\STATE{ sample $\{\textbf{x}, \textbf{y}\}$, a batch of real data.}
\STATE{ compute $\tilde{\textbf{x}} =\mathcal{G} (\textbf{x})$, a batch of generated data.}
\STATE compute $\ell_{1}= \mathcal{L}_{ce}(\mathcal{D}(\textbf{x}),\textbf{y})$
\IF{$\tau \text{ is defined}$}    
\STATE compute $\ell_{2}= \mathcal{L}_{ce}(\mathcal{D}(\tilde{\textbf{x}}), \tau)$ // Targeted
\ELSE
\STATE compute $\ell_{2}= \mathcal{L}_{ce}(\mathcal{D}(\tilde{\textbf{x}}), c^*-\textbf{y})$ // Untargeted
\ENDIF{}
\STATE update $\mathcal{D}'s$ weights to minimize $\lambda_1 \ell_{1} + \lambda_2\ell_{2} $
\STATE compute $\ell_3$=$L_1(\textbf{x},\tilde{\textbf{x}})$, pixel-wise loss between real and generated images.
\STATE update $\mathcal{G}'s$ weights to minimize $\gamma_1 \ell_{2} + \gamma_2\ell_3 $
\ENDWHILE
\RETURN $\mathcal{D},\mathcal{G}$
\end{algorithmic}
\end{algorithm}

\begin{figure*}
     \centering
     \begin{subfigure}[b]{0.42\textwidth}
         \centering
         \includegraphics[width=\textwidth]{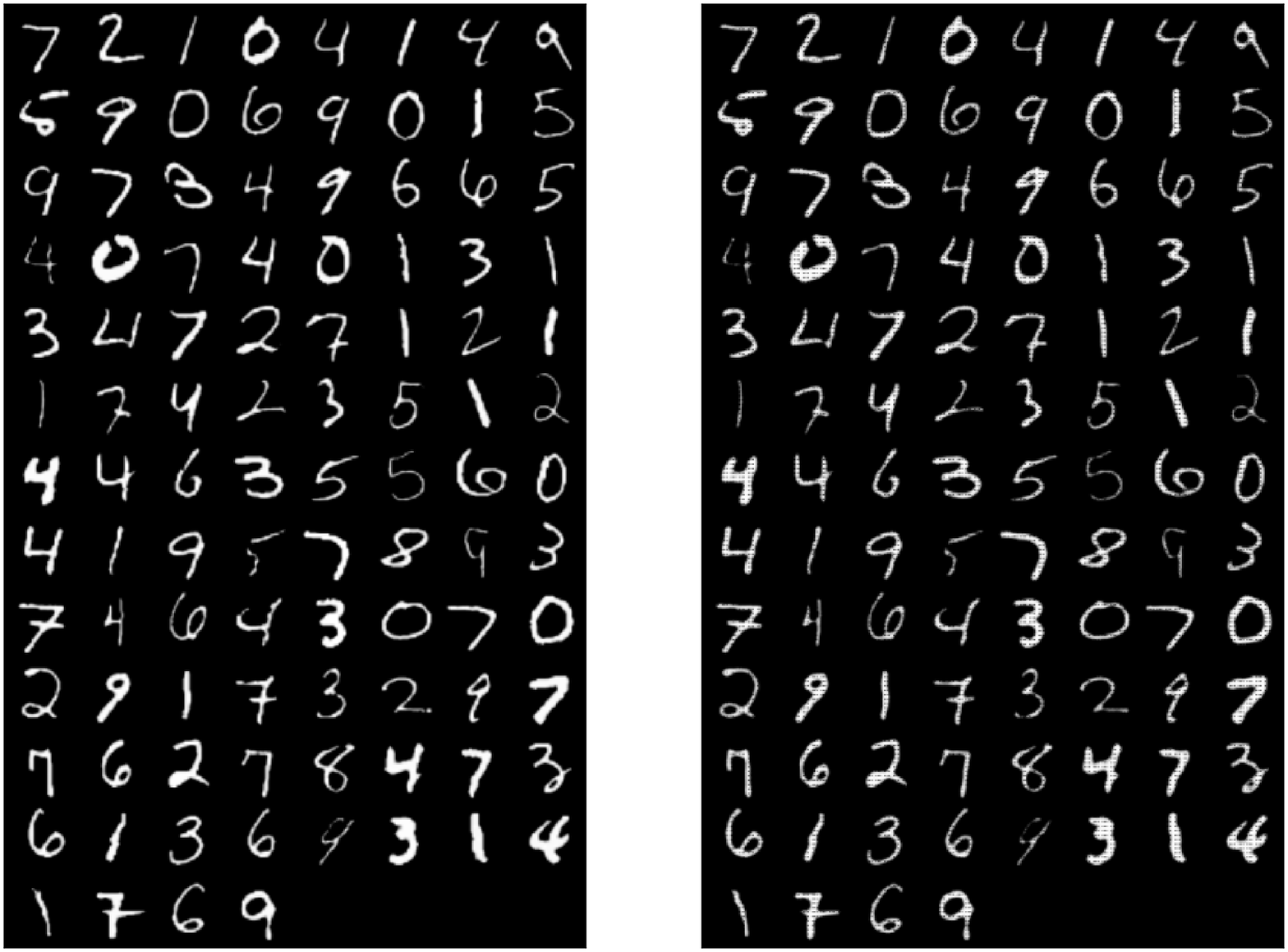}
         \caption{MNIST}
     \end{subfigure}
     \hfill
     \begin{subfigure}[b]{0.42\textwidth}
         \centering
         \includegraphics[width=\textwidth]{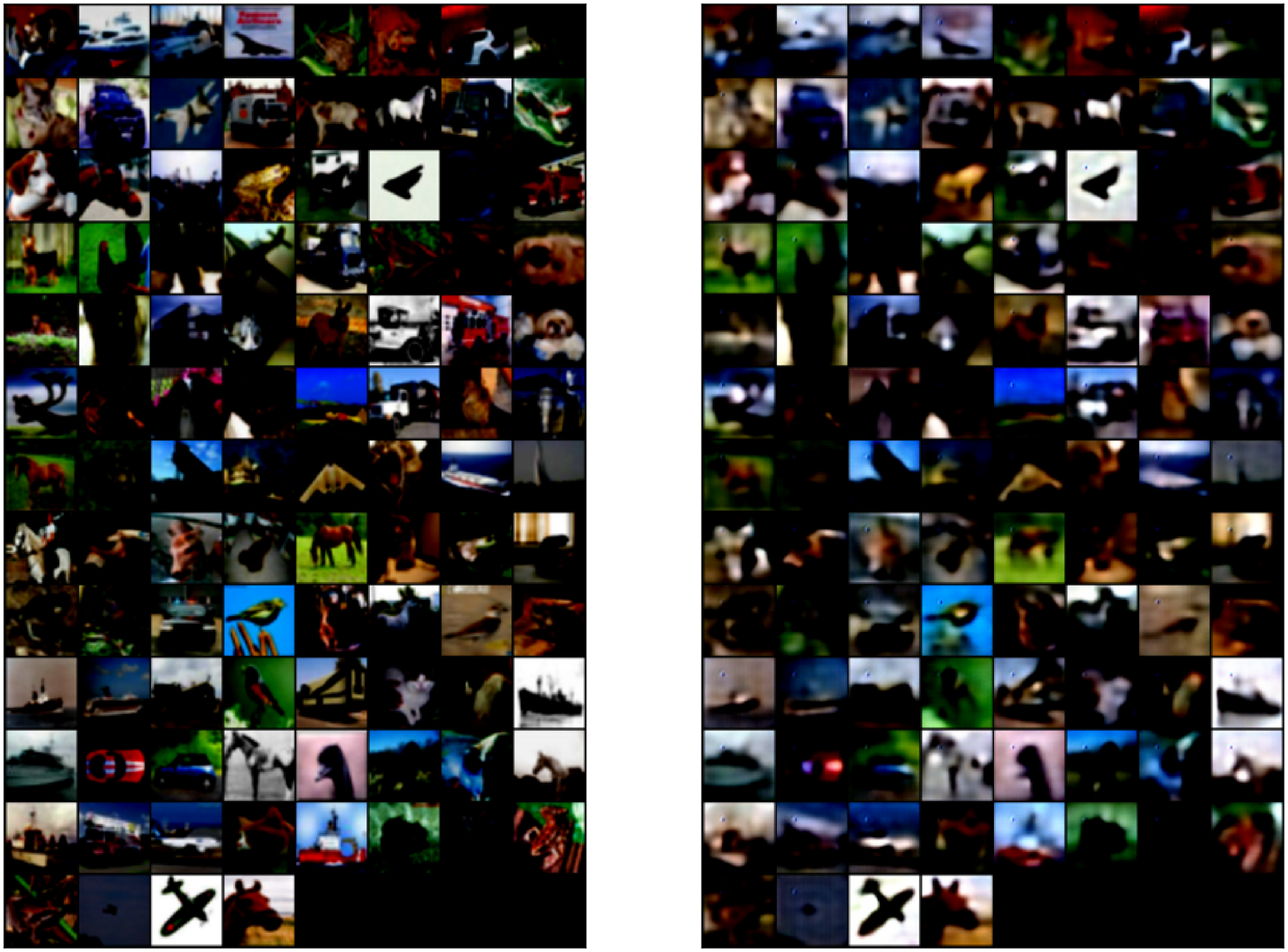}
         \caption{CIFAR10}
     \end{subfigure}
     \hfill
     \begin{subfigure}[b]{0.42\textwidth}
         \centering
         \includegraphics[width=\textwidth]{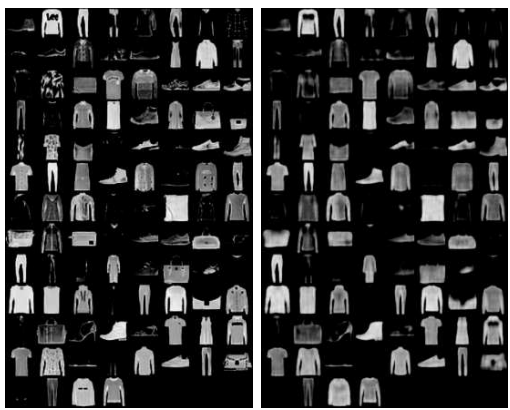}
         \caption{Fashion-MNIST}
     \end{subfigure}
     \\
     \begin{subfigure}[b]{0.42\textwidth}
         \centering
         \includegraphics[width=\textwidth]{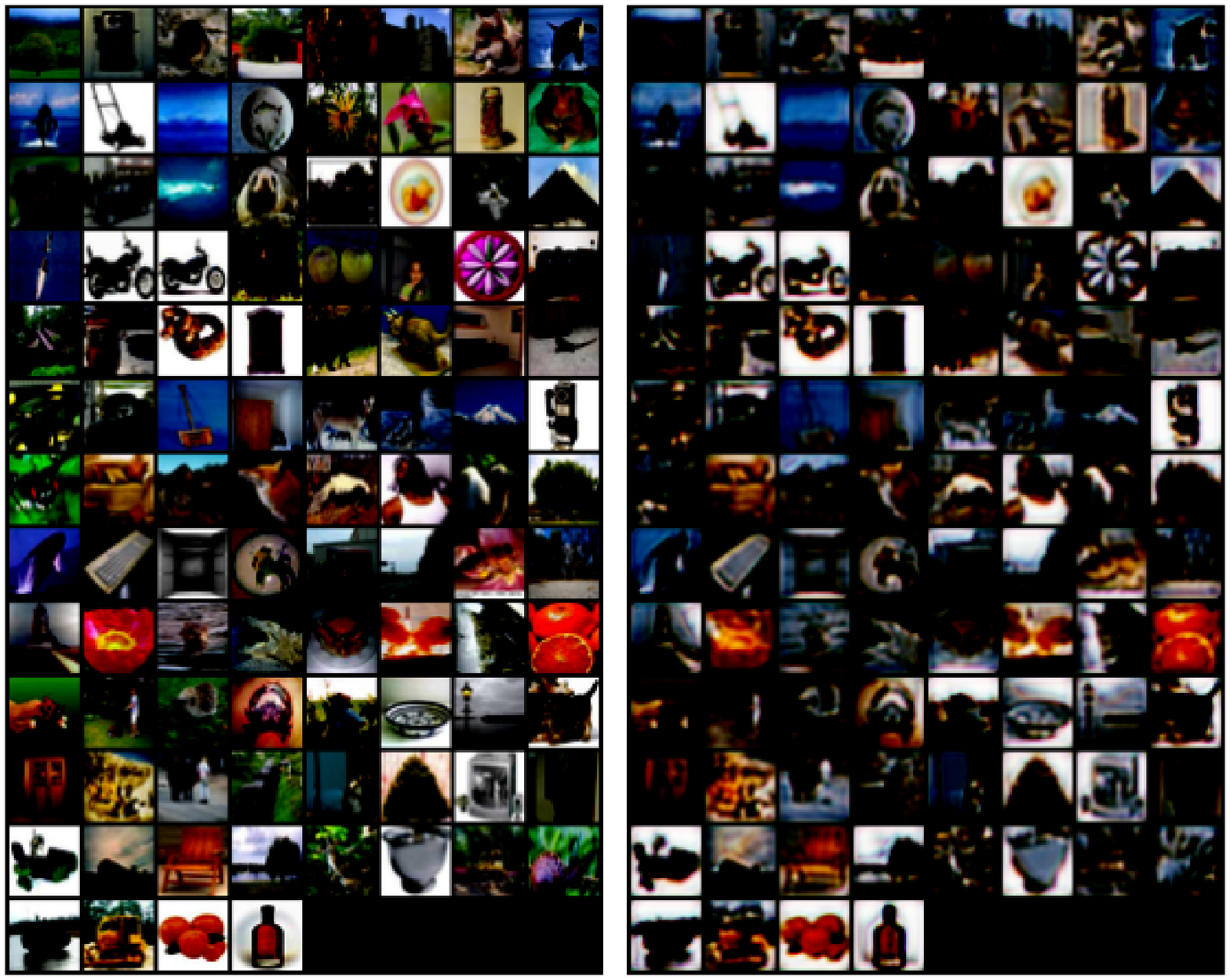}
         \caption{CIFAR100}
     \end{subfigure}
     \hfill
     \begin{subfigure}[b]{0.42\textwidth}
         \centering
         \includegraphics[width=\textwidth]{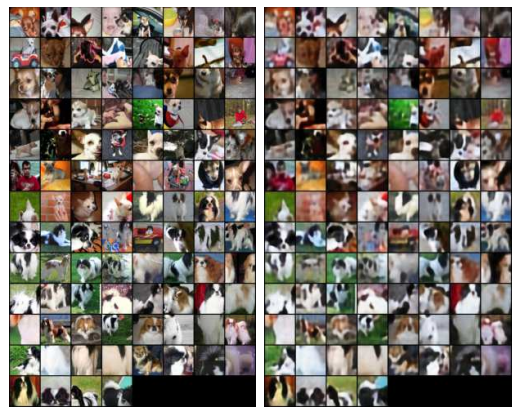}
         \caption{Stanford Dogs}
     \end{subfigure}
        \caption{Illustration of original images on the left and generated images on the right from (a) MNIST, (b) CIFAR10, (c) Fashion-MNIST, (d) CIFAR 100, and (e) Stanford Dogs datasets respectively, using the proposed method.}
        \label{fig:mnistA}
\end{figure*}

For the baseline network, we train a ResNet-18 based encoder and a custom decoder (3 transpose convolution layers interspersed with BatchNorm and ReLU). The encoder-decoder together serves as an autoencoder. We use the embedding produced by the encoder to reconstruct the perturbed data. The output of the last convolutional layer is set to $8$-D and $32$-D for MNIST and CIFAR10, respectively. The model is trained for $100$ epochs with both weight parameters in \autoref{eq:baseline_loss} set to $0.5$. After successful training, we extract the embedding of the training set per class to construct the hull of each class. Unfortunately, constructing a convex hull in higher dimensions is computationally infeasible, so we project the embedding onto 2D using PCA~\cite{Gewers_2022}. In each iteration, the data point moves towards the nearest face in the hull (or towards the centroid of target class). We then reconstruct the new point to its respective dimensions (8 and 32) using inverse PCA. The reconstructed point is then fed to the decoder and we observe the prediction by appending fully connected layers to the encoder. We perform this iteratively until we reach the terminating criterion.

\begin{figure}[h]
\includegraphics[width=7cm]{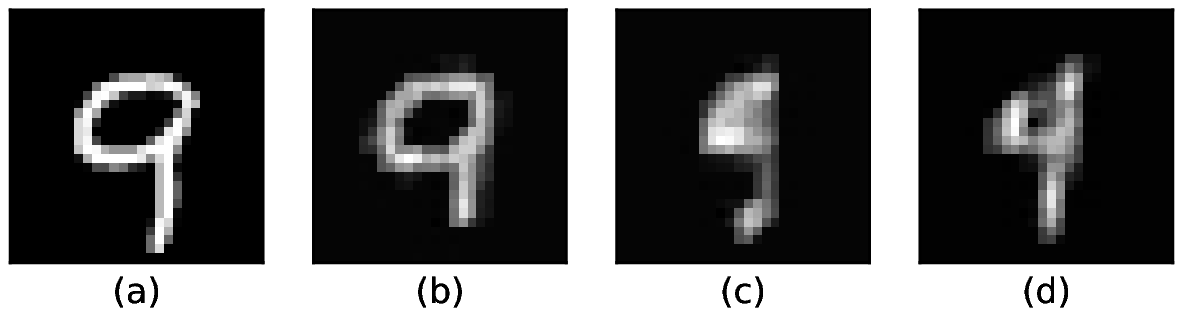}
\centering
\caption{Output of the baseline algorithm on an MNIST sample. (a) Original image. (b) Output of the autoencoder. (c) Reconstructed output of the decoder after the first iteration. (d) Reconstructed output of the decoder after convergence.}
\label{fig:orig_moved}
\end{figure}

\begin{table}[]
    \centering
    \begin{tabular}{cc}
       \includegraphics[scale=0.5]{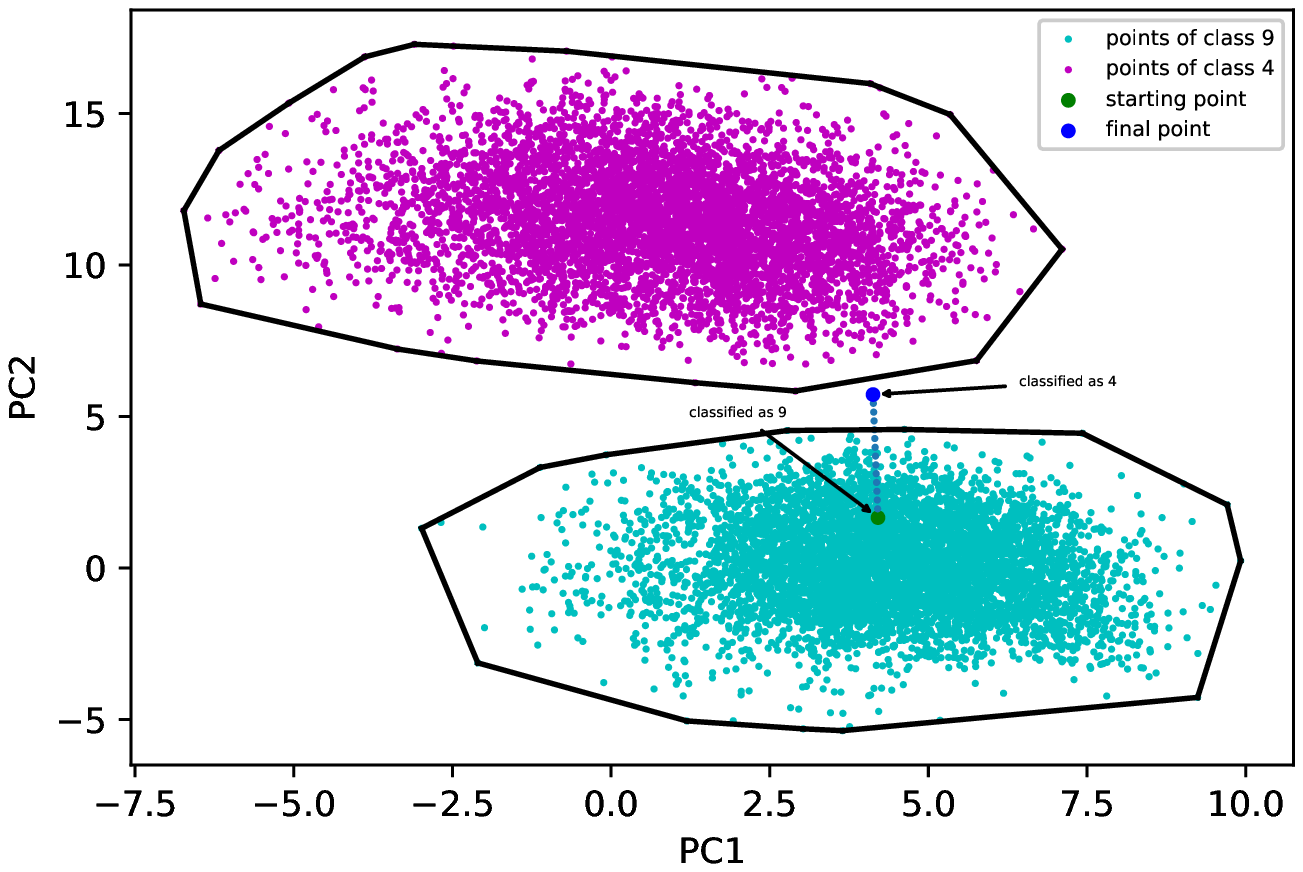}  & \includegraphics[scale=0.5]{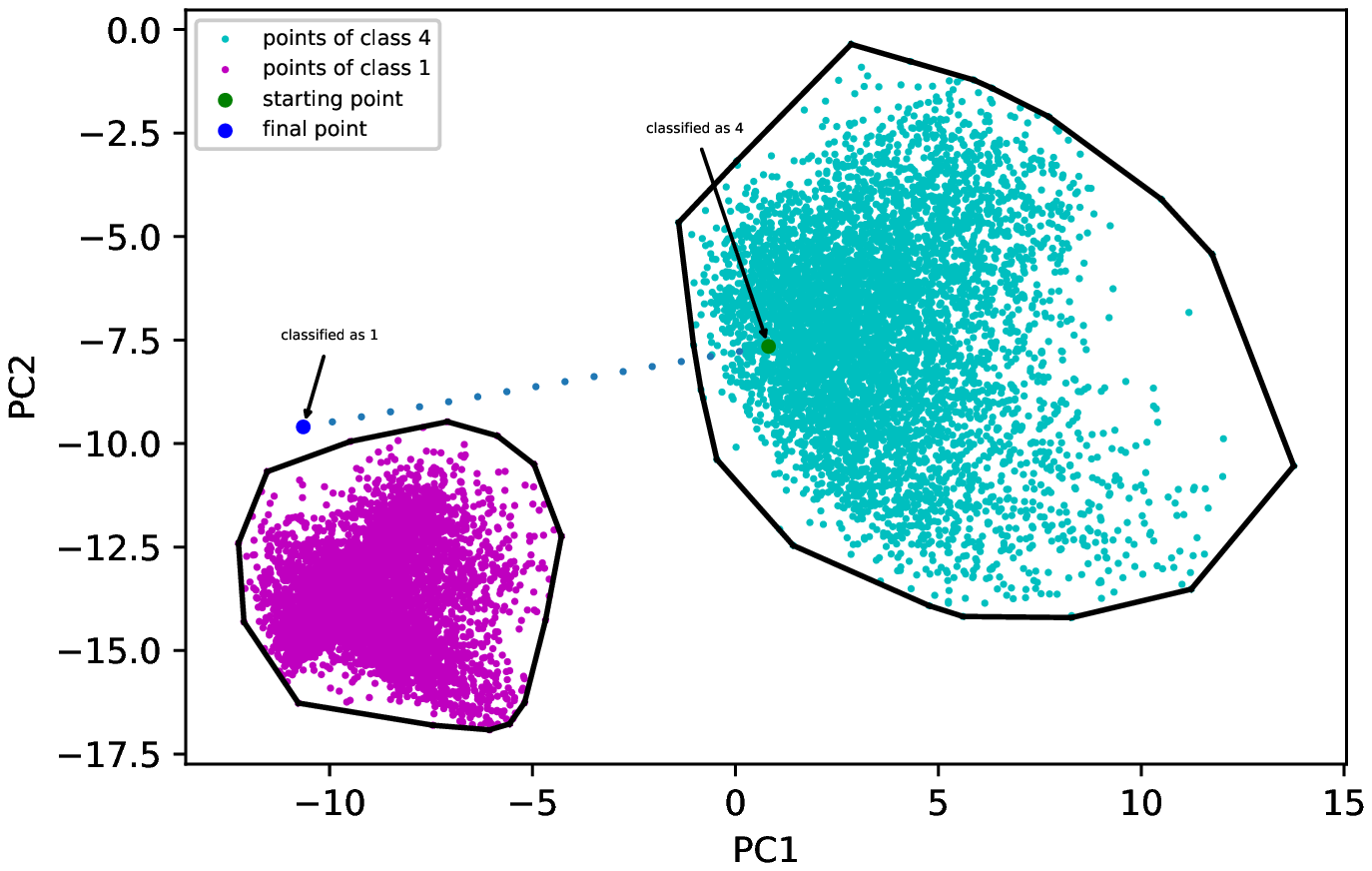} 
    \end{tabular}
    \caption{llustration of adversarial attacks from a geometric perspective using convex hulls. Left: Traversal from class `9' to target class `4' (targeted attack). Right: Traversal from class `4' towards the nearest class (untargeted attack).}
    \label{fig:trajectory}
\end{table}

\textbf{Experiments on MNIST and CIFAR10: } We conduct experiments on two standard datasets: MNIST~\cite{726791} and CIFAR10~\cite{Krizhevsky09learningmultiple}. We normalize and resize the images from both datasets to $64\times64$. We use the same train (50K) and test (10K) splits as defined in the datasets description page. We perform \textbf{both targeted and untargeted attacks} using the proposed algorithm. We quantify the success of the proposed method using the metric \textbf{attack success rate (ASR)} that evaluates the proportion of test data points that have been successfully (mis)classified as the target class in the case of targeted attack. For untargeted attack, ASR computes the proportion of test data points that have been classified as belonging to \textit{any} class other than their original class.   

\textbf{Experiments on Fashion-MNIST, CIFAR100 and Stanford Dogs: } 
To further validate the generalizability of the proposed method, we conduct experiments involving \textit{only} targeted attacks on three more datasets, namely, Fashion-MNIST~\cite{xiao2017/online}, CIFAR-100~\cite{cifar100} and Stanford Dogs~\cite{KhoslaYaoJayadevaprakashFeiFei_FGVC2011}. We use the same train and test splits provided by the original authors for each of the three datasets. For CIFAR100 and Stanford Dogs datasets, we randomly select 10 classes out of a total of 100 classes and 120 classes, respectively, and all the classes in Fashion-MNIST dataset to perform targeted attacks. 

\section{Findings}
\label{sec:find}

\subsection{Results}
\label{sec:result}


\begin{table*}[h]
\center
\caption{Adversarial success rates on MNIST on both targeted (left) and untargeted (right) attacks. We report the classification accuracy on original and generated images and ASR that denotes the attack success rate. Higher the value of ASR, the better is the performance.}
\scalebox{0.65}{
\begin{tabular}{|l|l|l|l|l|l|l||l|l|l|l|l|l|l|}
\hline
\multicolumn{7}{|c||}{Targeted Attacks}                                  & \multicolumn{7}{|c|}{Untargeted Attacks}                              \\ \hline
\multicolumn{1}{|l}{}\begin{tabular}{@{}l} Target \\ Class \end{tabular} &  \multicolumn{3}{|c|}{\multirow{1}{*}{Proposed Method}} & \multicolumn{3}{|c||}{\multirow{1}{*}{Baseline}}                                               & \multicolumn{1}{|l}{}\begin{tabular}{@{}l} Target \\ Class \end{tabular}  &  \multicolumn{3}{|c|}{\multirow{1}{*}{Proposed Method}} & \multicolumn{3}{|c|}{\multirow{1}{*}{Baseline}}                                           \\ \cline{2-7} \cline{9-14}                                                                                 & \begin{tabular}{@{}l}Acc. on\\ Original\\ Images\end{tabular}  & \begin{tabular}{@{}l}Acc. on\\ Generated\\ Images\end{tabular}   & ASR  &  \begin{tabular}{@{}l}Acc. on\\ Original\\ Images\end{tabular} & \begin{tabular}{@{}l}Acc. on\\ Generated\\ Images\end{tabular} & ASR &                                                                          & \begin{tabular}{@{}l}Acc. on\\ Original\\ Images\end{tabular}                                                                                        & \begin{tabular}{@{}l}Acc. on \\ Generated\\ Images\end{tabular}  & ASR  & \begin{tabular}{@{}l}Acc. on\\ Original\\ Images\end{tabular} & \begin{tabular}{@{}l}Acc. on\\ Generated \\ Images\end{tabular} & ASR \\ \hline \hline
0 & 99.06\% & 3.00\% &97.00\% & 99.40\%&81.87\%  &  18.13\%  &0 & 60.61\% & 0.00\% & 100.00\% & 99.40\%& 6.84\% & 93.16\% \\

1 & 99.63\% &7.00\% &93.00\% & 99.30\%&99.93\%    & 0.07\%  &1 & 93.57\% & 0.00\% & 100.00\% & 99.30\%& 97.44\% & 2.56\%\\

2 & 96.13\% &3.95\% &96.05\% & 98.40\%&37.05\%   & 62.95\% & 2 & 81.01\% & 16.86\% & 83.14\% & 98.40\%& 0.00\% &100.00\% \\

3 & 86.74\% &5.73\% &94.27\% & 99.50\%&50.86\%  & 49.14\%& 3 & 93.37\% & 1.58\% & 98.42\% & 99.50\%&99.60\% &0.40\% \\

4 & 83.06\% &3.48\%  &96.52\% & 99.20\%& 57.26\% & 42.74\% &4 & 88.09\% & 18.02\% & 81.98\% & 99.20\%&1.93\%  &98.07\% \\

5 & 99.77\% &22.31\%  &77.69\% & 99.30\%& 82.48\%  & 17.52\%& 5 & 92.49\% & 0.67\% & 99.33\% & 99.30\%& 12.67\%& 87.33\%\\

6 & 54.14\% &1.56\%  &98.44\% & 98.20\%& 40.31\%  & 59.69\%&6 & 90.50\% & 45.51\% & 54.49\% & 98.20\%& 6.26\% &93.74\% \\

7 & 91.56\% &3.76\% &96.24\% & 98.30\%& 27.50\%  & 72.50\% & 7 & 88.52\% & 0.19\% & 99.81\% & 98.30\%& 89.59\% &10.41\% \\

8 & 88.21\% &1.83\%  &98.17\% & 98.80\%& 60.95\%  & 39.05\%& 8 & 81.52\% & 8.73\% & 91.27\% & 98.80\%& 95.28\% &4.72\% \\

9 & 89.72\% &9.76\% &90.24\% & 98.30\%& 9.54\%  & 32.06\%& 9 & 91.38\% & 0.10\% & 99.90\% & 98.30\%& 97.03\% & 2.97\%
\\ \hline
\textbf{Avg.} & \textbf{88.80\%} &\textbf{6.23\%}&\textbf{92.86\%} &\textbf{98.88\%} & \textbf{67.94\%}  & \textbf{32.06\%}&
\textbf{Avg.} & \textbf{86.18\%} &\textbf{9.10\%}&\textbf{90.83\%} &\textbf{98.88\%} & \textbf{50.60\%}  & \textbf{49.40\%}
\\ \hline
\end{tabular}}
\label{tab:mnist_acc}
\end{table*}

\begin{table*}[h]
\center
\caption{Adversarial success rates on CIFAR10 on both targeted (left) and untargeted attacks (right). We report the classification accuracy on original and generated images and ASR that denotes the attack success rate. Higher the value of ASR, the better is the performance.}
\scalebox{0.65}{
\begin{tabular}{|l|l|l|l|l|l|l||l|l|l|l|l|l|l|}
\hline
\multicolumn{7}{|c||}{Targeted Attacks}                                  & \multicolumn{7}{|c|}{Untargeted Attacks}                              \\ \hline
\multicolumn{1}{|l}{}\begin{tabular}{@{}l} Target \\ Class \end{tabular} &  \multicolumn{3}{|c|}{\multirow{1}{*}{Proposed Method}} & \multicolumn{3}{|c||}{\multirow{1}{*}{Baseline}}                                               & \multicolumn{1}{|l}{}\begin{tabular}{@{}l} Target \\ Class \end{tabular}  &  \multicolumn{3}{|c|}{\multirow{1}{*}{Proposed Method}} & \multicolumn{3}{|c|}{\multirow{1}{*}{Baseline}}                                           \\ \cline{2-7} \cline{9-14}                                                                                 & \begin{tabular}{@{}l}Acc. on\\ Original\\ Images\end{tabular}  & \begin{tabular}{@{}l}Acc. on\\ Generated\\ Images\end{tabular}   & ASR  &  \begin{tabular}{@{}l}Acc. on\\ Original\\ Images\end{tabular} & \begin{tabular}{@{}l}Acc. on\\ Generated\\ Images\end{tabular} & ASR &                                                                          & \begin{tabular}{@{}l}Acc. on\\ Original\\ Images\end{tabular}                                                                                        & \begin{tabular}{@{}l}Acc. on \\ Generated\\ Images\end{tabular}  & ASR  & \begin{tabular}{@{}l}Acc. on\\ Original\\ Images\end{tabular} & \begin{tabular}{@{}l}Acc. on\\ Generated \\ Images\end{tabular} & ASR \\ \hline \hline
0 & 75.00\% &37.40 &62.60\% & 94.70\%&  10.61\% &  0.00\%  &0 & 87.20\% & 2.80\% & 97.20\% & 94.70\%& 0.00\%  &100.00\%  \\ 
1 & 47.67\% &7.60\% & 92.40\%  & 97.30\%& 10.61\%    & 0.00\% &1 & 62.00\% & 0.00\% & 100.00\% & 97.30\%& 0.00\%&100.00\%  \\ 
2 & 76.43\% &6.41\% &93.59\% & 90.10\%&10.10\%   &19.65 \% & 2 & 78.60\% & 0.60\% & 99.40\% & 90.10\%& 7.40\% & 92.60\% \\  
3 & 74.73\% &6.12\% &93.88\% & 87.50\%&9.97\%  & 97.50\% &3 & 66.40\% & 65.10\% & 34.90\% & 87.50\%& 58.80\% & 41.20\% \\  
4 & 60.68\%&5.65\% &94.35\% & 95.00\%&9.99\% & 0.41\%  & 4 & 97.90\% & 11.60\% & 88.40\% & 95.00\%&  16.10\%& 83.90\% \\ 
5 & 77.51\%&9.41\%  &90.59\% & 89.00\%& 10.00\%  & 0.00\% & 5 & 68.00\% & 29.90\% & 70.10\% & 89.00\%& 0.00\% & 100.00\% \\ 
6 & 65.57\% &18.60\%   &81.40\% & 95.30\%& 10.39\%  & 0.00\% &6 & 95.80\% & 1.90\% & 98.10\% & 95.30\%&0.00\%  & 100.00\% \\ 
7 & 76.5 \%&6.86\%  &93.14\% & 94.70\%& 10.08\%  & 0.00\% &7 & 84.70\% & 1.60\% & 98.40\% & 94.70\%& 0.30\% &99.70\%  \\
8 &  76.35\%&16.78\% &83.22\% & 96.00\%& 10.58\%  & 38.92\% & 8 & 92.60\% & 11.40\% & 88.60\% & 96.00\%& 1.00\% &99.00\% \\   
9 & 87.24\%&27.68\% &72.32\% & 97.00\%&10.00\%  & 0.00\% &9 & 69.50\% & 0.00\% & 100.00\% & 97.00\%& 0.00\% &100.00 \% \\ 
\hline
\textbf{Avg.} & \textbf{71.70\%} &\textbf{14.26\%}&\textbf{85.74\%}& \textbf{93.66\%}  & \textbf{10.20\%} &  \textbf{15.64\%} &
\textbf{Avg.} & \textbf{80.20\%} &\textbf{12.40\%}&\textbf{87.51\%} &\textbf{93.66\%} & \textbf{ 0.83\%}  & \textbf{91.60\%}
\\ \hline
\end{tabular}}
\label{tab:cifar_acc}
\end{table*}

\begin{table*}[h]
\centering
\caption{Adversarial success rates on Fashion-MNIST, CIFAR100 and Stanford Dogs on \textit{only} targeted attacks using the proposed method. We report the classification accuracy on original and generated images, and ASR that denotes the attack success rate. Higher the value of ASR, better is the performance.}
\scalebox{0.8}{
\begin{tabular}{|l|l|l|l|l|l|l|l|l|l|l|l|}
\hline
\multicolumn{4}{|c|}{Fashion-MNIST}    &                              
\multicolumn{4}{c|}{CIFAR100}         & 
\multicolumn{4}{c|}{StanfordDogs}  \\
\hline
\begin{tabular}[|l|]{@{}l@{}}Target \\ Class \end{tabular} &
\begin{tabular}{@{}l}Acc. on\\ Original\\ Images\end{tabular}  & \begin{tabular}{@{}l}Acc. on\\ Generated\\ Images\end{tabular}   & ASR    
&
\begin{tabular}[|l|]{@{}l@{}}Target \\ Class \end{tabular} &
\begin{tabular}{@{}l}Acc. on\\ Original\\ Images\end{tabular}  & \begin{tabular}{@{}l}Acc. on\\ Generated\\ Images\end{tabular}   & ASR    
&
\begin{tabular}[|l|]{@{}l@{}}Target \\ Class \end{tabular} &
\begin{tabular}{@{}l}Acc. on\\ Original\\ Images\end{tabular}  & \begin{tabular}{@{}l}Acc. on\\ Generated\\ Images\end{tabular}   & ASR  

\\
 \hline
0 & 84.78\% & 2.87\% & 97.13\%     & 6&53.33\%  &  22.43\% & 77.57\%         & 23 & 22.11\% & 1.32\%&99.68\%\\

1 & 89.45\% & 19.60\% &80.40\%      & 43&57.87\%  &  0.56\% & 99.44\% 
& 0 & 12.43\% & 0.55\%& 99.45\% \\

2 & 91.45\% & 1.37\% &98.63\%      & 12&52.10\%  &  13.54\%  & 86.46\% 
& 12 & 21.65\% & 1.03\% & 98.97\% \\

3 & 89.43\% & 4.32\% &95.68\%       & 76&58.75\%  &  1.56\%   & 98.44\% 
& 46 & 20.76\% & 0.87\% & 99.13\% \\

4 & 91.11\% & 3.93\% & 96.07\%       & 34&55.33\%  &  6.76\%   & 93.24\% 
& 55 & 29.65\% & 1.49\% & 99.51\% \\

5 & 88.45\% & 2.84\% &97.16\%      &  91&49.63\%  &  3.56\%   & 96.44\% 
& 81 & 19.55\% & 1.34\% & 98.66\% \\

6 & 88.44\% & 2.59\% &97.41\%      & 11&55.91\%  &  5.51\%   & 94.49\% 
& 92 & 10.10\% & 0.98\% & 99.02\% \\

7 & 92.86\% & 10.98\% &89.02\%      & 71&57.60\%  &  2.93\%   & 97.07\% 
& 60 & 20.98\% & 1.01\% & 98.99\% \\

8 & 91.11\% & 3.23\% &96.77\%      & 23&49.78\%  &  1.76\%  & 98.24\% 
& 16 & 23.44\% & 0.87\% & 99.13\% \\

9 & 90.24\% & 11.72\% &88.28\%      & 10&50.54\%  &  3.61\%   & 96.39\% 
& 39 & 22.25\% & 1.19\% & 99.81\%

\\ \hline
\textbf{Avg.} & \textbf{89.83\%} & \textbf{6.34\%} & \textbf{93.66\%} &
\textbf{Avg.} & \textbf{54.28\%}  & \textbf{6.22\%} & \textbf{93.78\%} & 
\textbf{Avg.} & \textbf{20.29\%} & \textbf{1.06\%} & \textbf{98.94\%} \\ 
\hline
\end{tabular}}
\label{tab:stanford_acc}
\end{table*}

\noindent\textbf{Proposed Method Results: } We present the results of our proposed method on MNIST, CIFAR10, Fashion-MNIST, CIFAR100 and Stanford Dogs datasets in \autoref{fig:mnistA}.  On MNIST (see \autoref{tab:mnist_acc}), we observe classification accuracy of 88.8\% on original images, 10.5\% on perturbed images and secure an attack success rate of 92.86\% when averaged across all the classes for targeted attacks. We observe classification accuracy of 86.18\% on original images, 9.1\% on perturbed images and secure an attack success rate of 90.83\% when averaged across all the classes for untargeted attacks. On CIFAR10 (see \autoref{tab:cifar_acc}), we observe classification accuracy of 71.7\% on original images, 10.01\% on perturbed images and secure an attack success rate of 85.74\% when averaged across all the classes for targeted attacks. We observe classification accuracy of 80.20\% on original images, 12.4\% on perturbed images and secure an attack success rate of 87.51\% when averaged across all the classes for untargeted attacks. We also test our method on Fashion-MNIST, CIFAR100 and Stanford Dogs datasets on targeted attacks to validate the generalizability of the proposed method. We randomly select 10 classes out of a total of 100 classes from the CIFAR100 and out of 120 classes from the Stanford Dogs datasets, respectively. We report the results in \autoref{tab:stanford_acc}. We observe that the proposed method achieves an ASR of 93.66\% on the Fashion-MNIST dataset, an ASR of 93.78\% on the CIFAR100 dataset and an ASR of 98.94\% on the Stanford Dogs dataset. We suspect that a low classification accurccy on original images from the Stanford Dogs dataset may be responsible for surprisingly high value of ASR.


 To quantify the preservation of \textit{visual realism} between original and generated images, we use two measures: 1) \textbf{Structural Similarity Index Measure (SSIM)} 2) \textbf{Peak Signal-to-Noise Ratio (PSNR)}. We use the MNIST dataset for this purpose. We report SSIM = 0.85 for untargeted attacks and SSIM = 0.78 for targeted attacks. We observe PSNR = 21.7 for untargeted attacks and PSNR = 17.1 for targeted attacks.

 \begin{figure}
     \centering
     \begin{subfigure}[b]{0.18\textwidth}
         \centering
         \includegraphics[width=\textwidth]{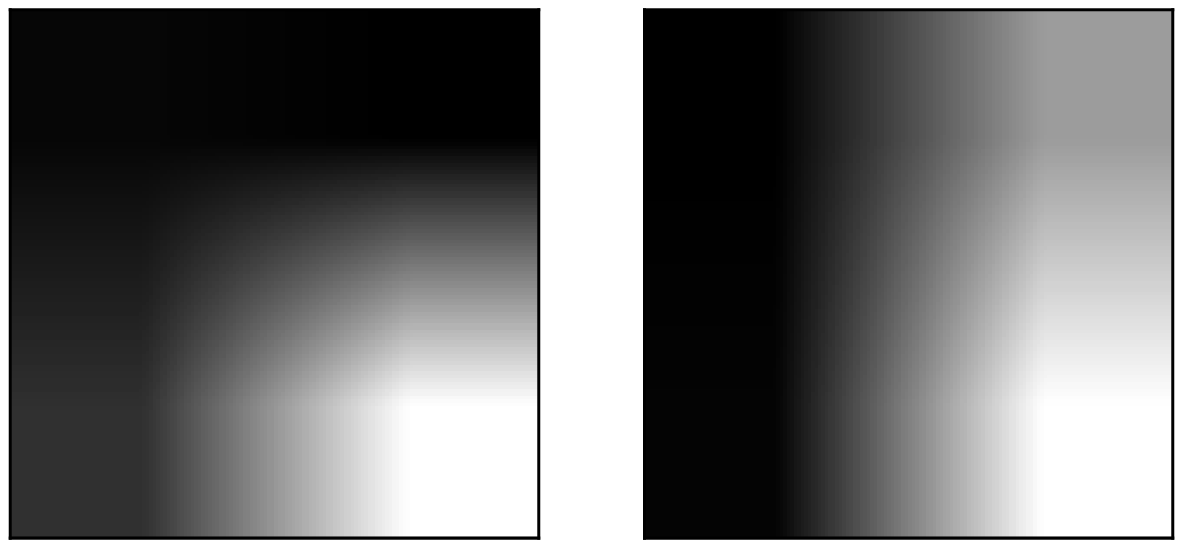}
         \caption{Sample \#1}
     \end{subfigure}  \hspace{0.3cm}
     \begin{subfigure}[b]{0.18\textwidth}
         \centering
         \includegraphics[width=\textwidth]{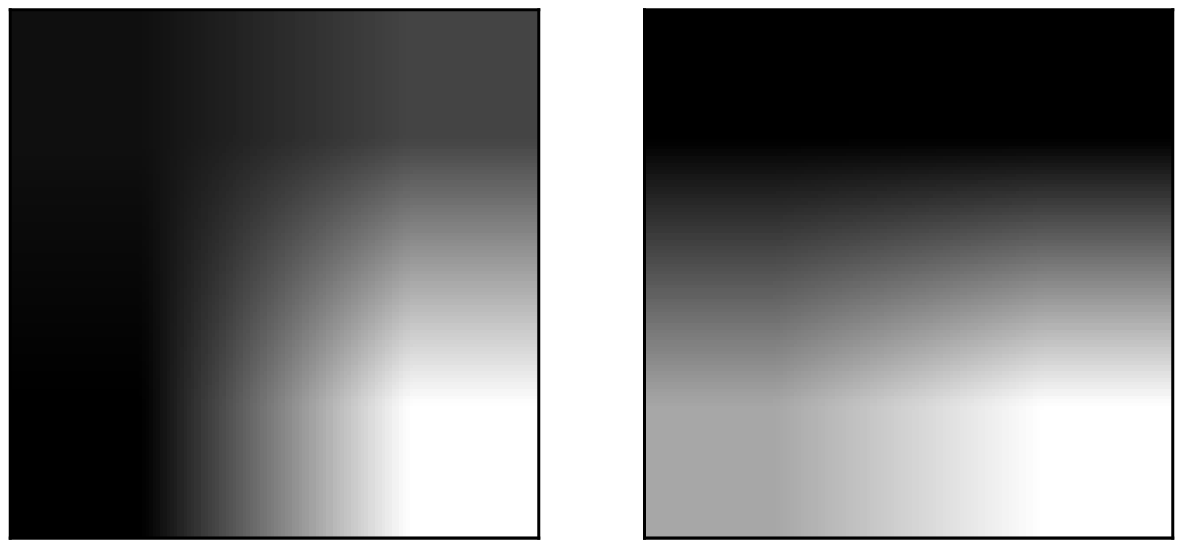}
         \caption{Sample \#17}
     \end{subfigure}
     \\
     \begin{subfigure}[b]{0.18\textwidth}
         \centering
         \includegraphics[width=\textwidth]{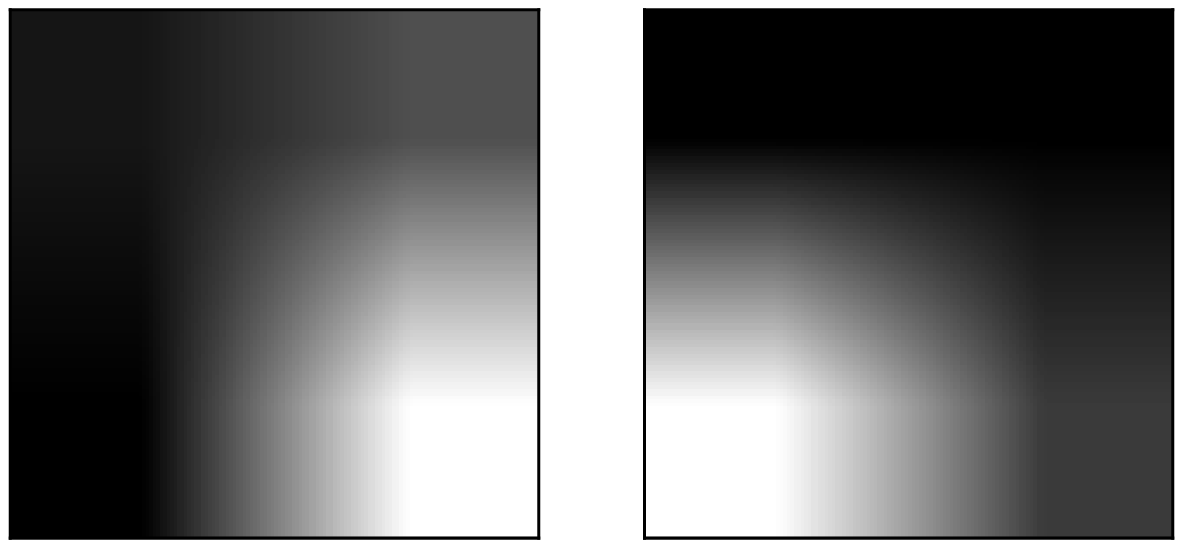}
         \caption{Sample \#5}
     \end{subfigure}    \hspace{0.3cm} 
     \begin{subfigure}[b]{0.18\textwidth}
         \centering
         \includegraphics[width=\textwidth]{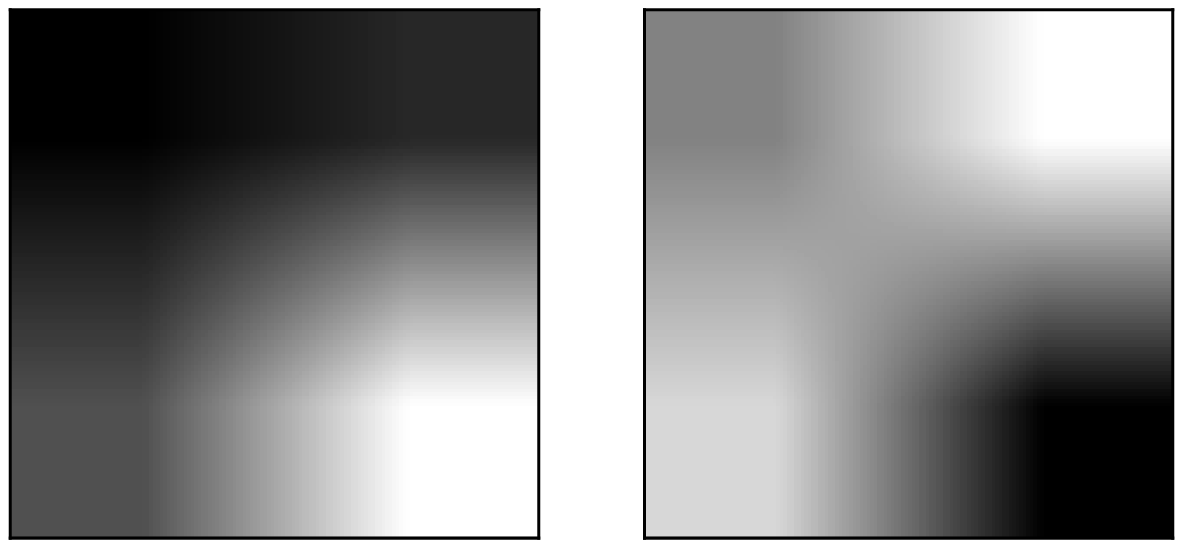}
         \caption{Sample \#8}
     \end{subfigure}
        \caption{Activation maps corresponding to class predicted by $\mathcal{D}$ (left)  and target class (right) for four different test images belonging to class `0' in the CIFAR10 dataset.}
        \label{fig:gradcam}
\end{figure}

\noindent\textbf{Baseline Results: } \autoref{fig:orig_moved} depicts perturbation in the latent space to modify the class label of the reconstructed image changes without affecting its perceptual quality using the autoencoder. The objective is to move towards a different class than the original class in the case of untargeted attacks. In the case of targeted attacks, we migrate towards the centroid of the convex hull of the target class. We offer a geometrical interpretation of how the baseline routine works in \autoref{fig:trajectory}. We demonstrate how the addition of adversarial perturbation in the latent space pushes the data point from the convex hull spanned by the training data points belonging to the `original' class  towards the hull of the `target' class resulting in an adversarial attack. Given a test sample $x$ of class $c$, lying in the hull constructed by features of training samples in that class, we add iterative perturbations to $f(x)$ towards the direction of the nearest face in the hull until the classification of the reconstructed image changes. In the case of targeted attack, we move towards the centroid of the target class's hull. 
On MNIST (see \autoref{tab:mnist_acc}), we observe classification accuracy of 98.88\% on original images, 9.76\% on perturbed images and secure an attack success rate of 32.06\% when averaged across all the classes for targeted attacks. We observe classification accuracy of 50.60\% on perturbed images and secure an attack success rate of 49.40\% when averaged across all the classes for untargeted attacks. On CIFAR10 (see \autoref{tab:cifar_acc}), we observe classification accuracy of 93.66\% on original images, 10.20\% on perturbed images and secure an attack success rate of 15.64\% when averaged across all the classes for targeted attacks. We observe classification accuracy of 0.83\% on perturbed images and secure an attack success rate of 91.60\% when averaged across all the classes for untargeted attacks. On CIFAR10, the baseline achieves a high ASR due to poor quality of generated images. We computed the structural similarity index measure (SSIM) between the original and generated images (baseline) and observed it to be very low $\sim$1\%. Due to the low quality of generated images, the classifier misclassifies the generated data to any arbitrary class leading to a spurious increase in the ASR. In all the remaining cases, the proposed method outperforms the baseline by a significant margin.

We further compare with two gradient-based attacks in the \textit{pixel space}, \textbf{FGSM} and \textbf{PGD} (both depend on noise margin $\epsilon$)~\cite{Review1} in Table~\ref{PGD} in terms of classification accuracy, and observe that the proposed method is agnostic to noise $\epsilon$ and outperforms FGSM but is comparable to PGD attacks in the pixel space on adversarially generated images.
 Additionally, we compare with \textbf{ATGAN} (attack without target model using GAN)~\cite{AWTGAN} and observe that ATGAN achieves the maximum attack success rate (ASR) of 81.78\% on the MNIST dataset (proposed method achieves 90.83\% ASR), and an ASR of 87.99\% on the CIFAR10 dataset (proposed method achieves 87.51\%).
We also compare the $L_2$ norm of perturbation added by the proposed method. On MNIST, we observe the average $L_2$ norm of noise in pixel space to be $0.079$ and $2.559$ in $\mathcal{D}'s$ latent space. On CIFAR 10, these values are $0.016$ and $0.519$, respectively.

\begin{figure*}[t]
     \centering
     \begin{subfigure}[b]{0.3\textwidth}
         \centering
         \includegraphics[width=\textwidth]{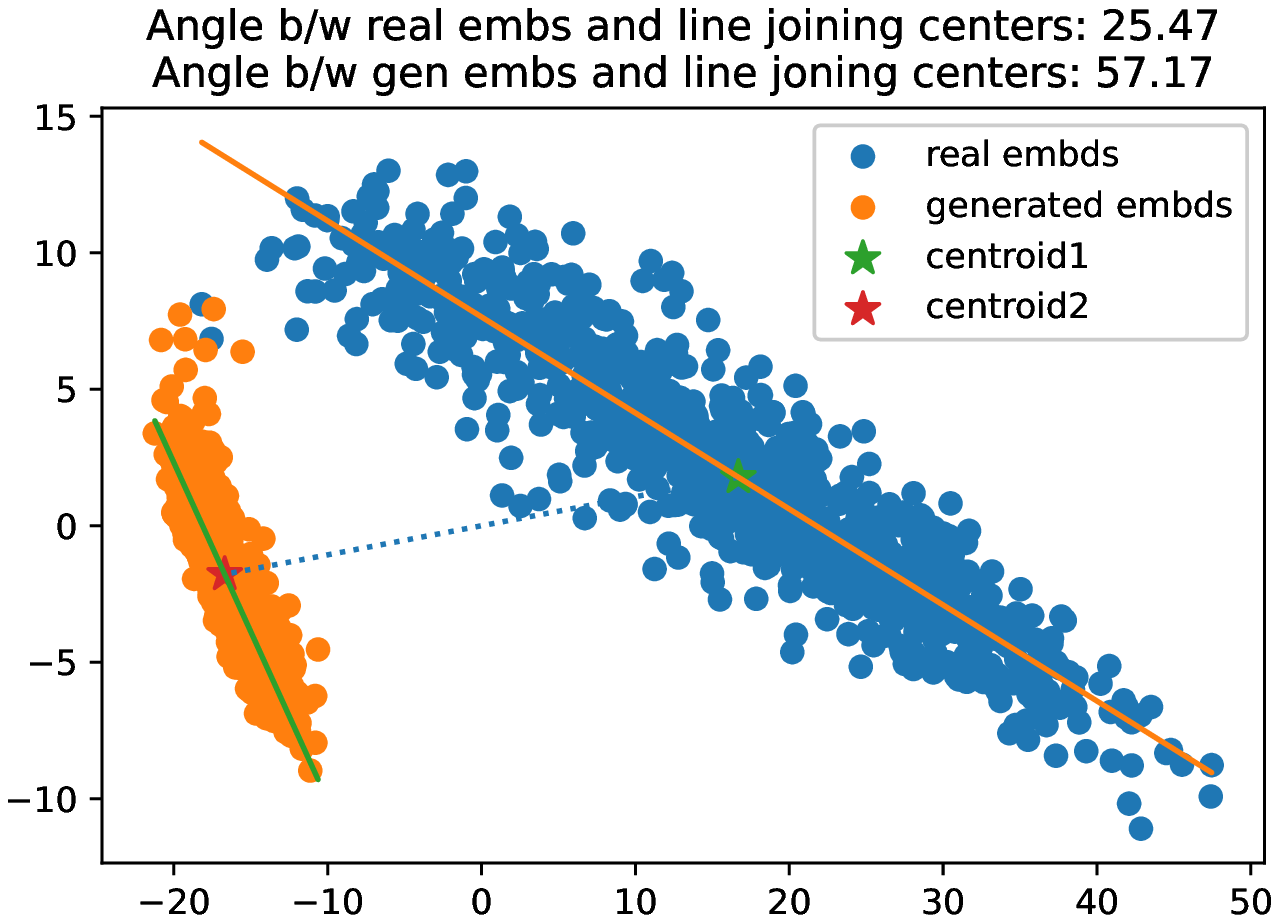}
     \end{subfigure}  
     \begin{subfigure}[b]{0.3\textwidth}
         \centering
         \includegraphics[width=\textwidth]{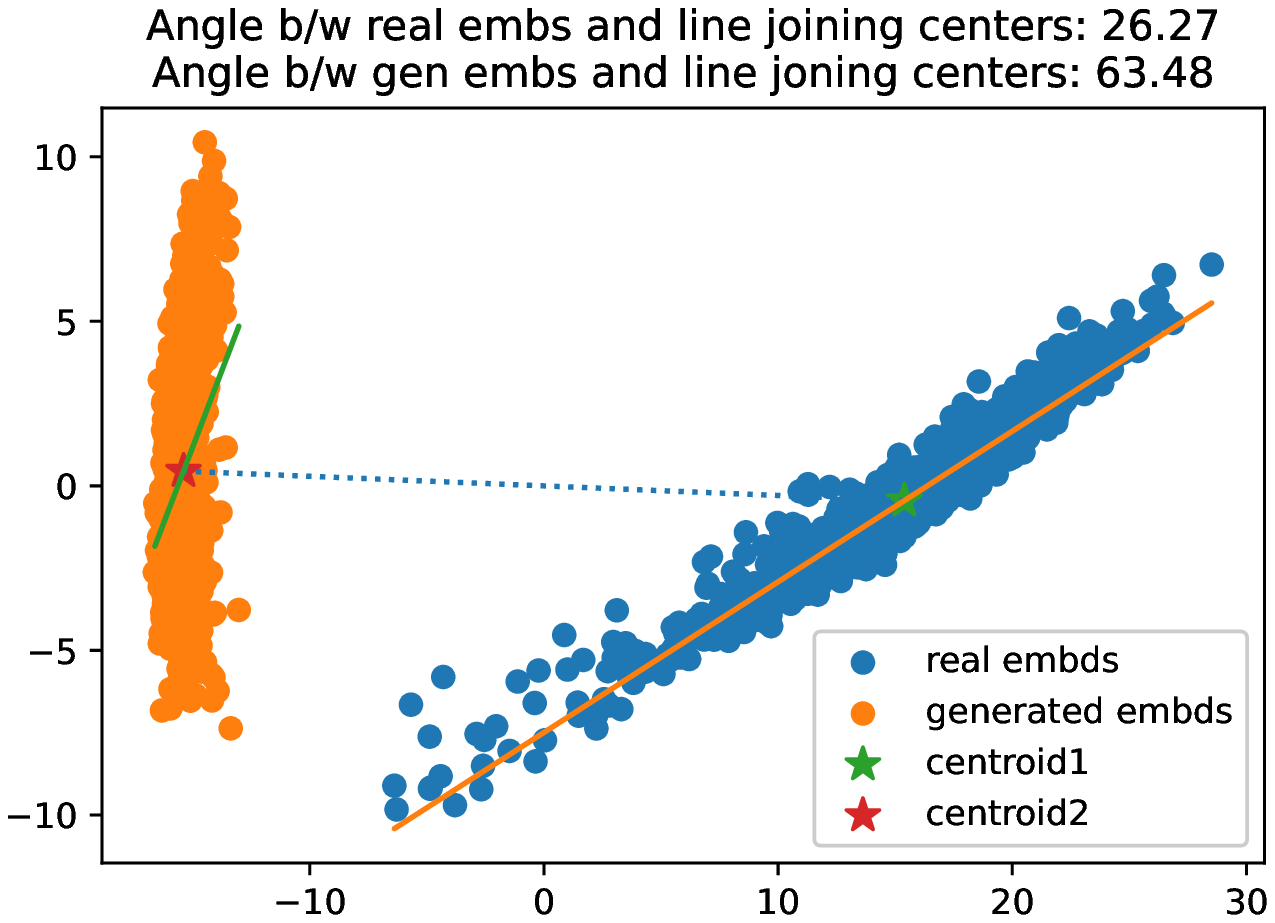}
     \end{subfigure}
     \begin{subfigure}[b]{0.3\textwidth}
         \centering
         \includegraphics[width=\textwidth]{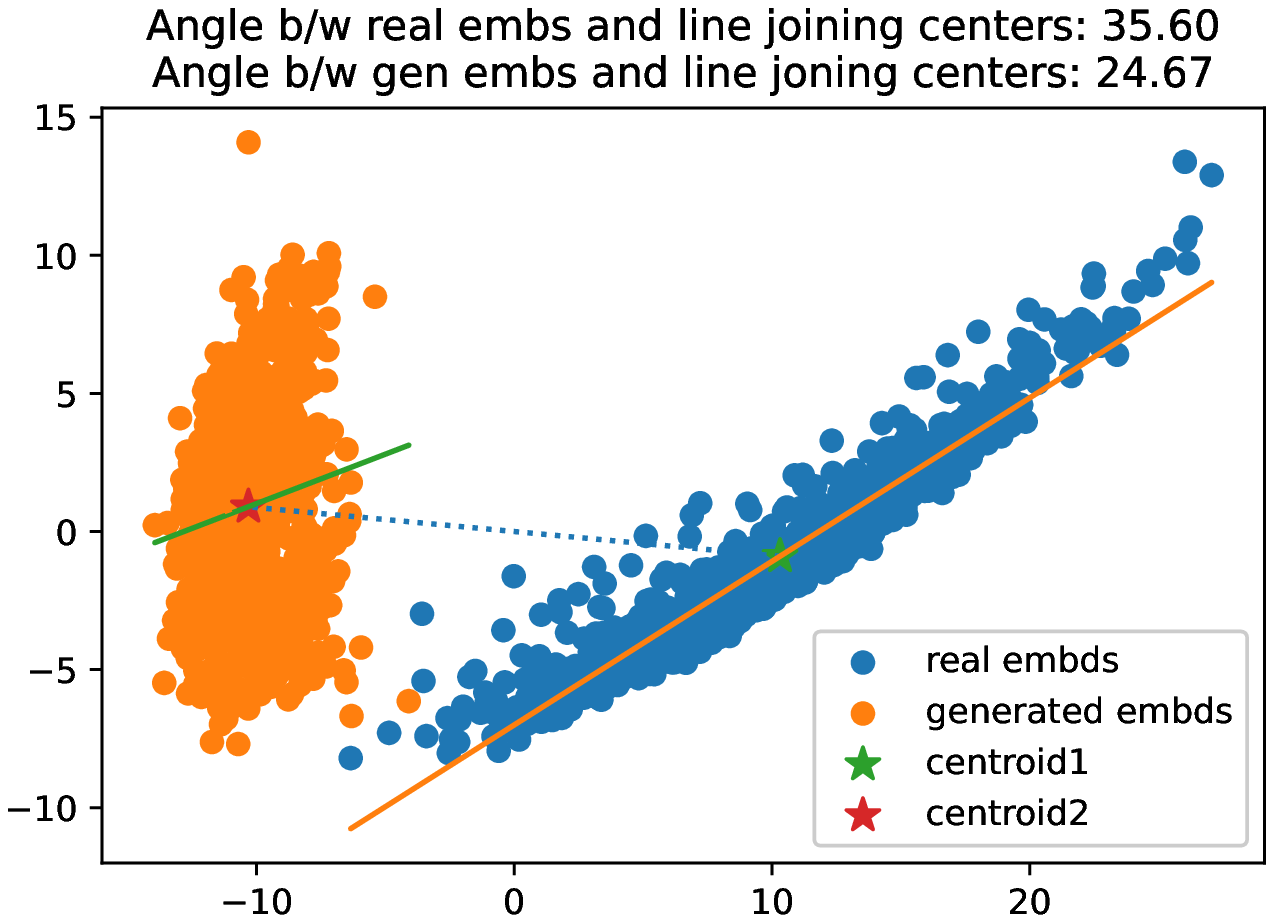}
     \end{subfigure}  \\  
     \begin{subfigure}[b]{0.3\textwidth}
         \centering
         \includegraphics[width=\textwidth]{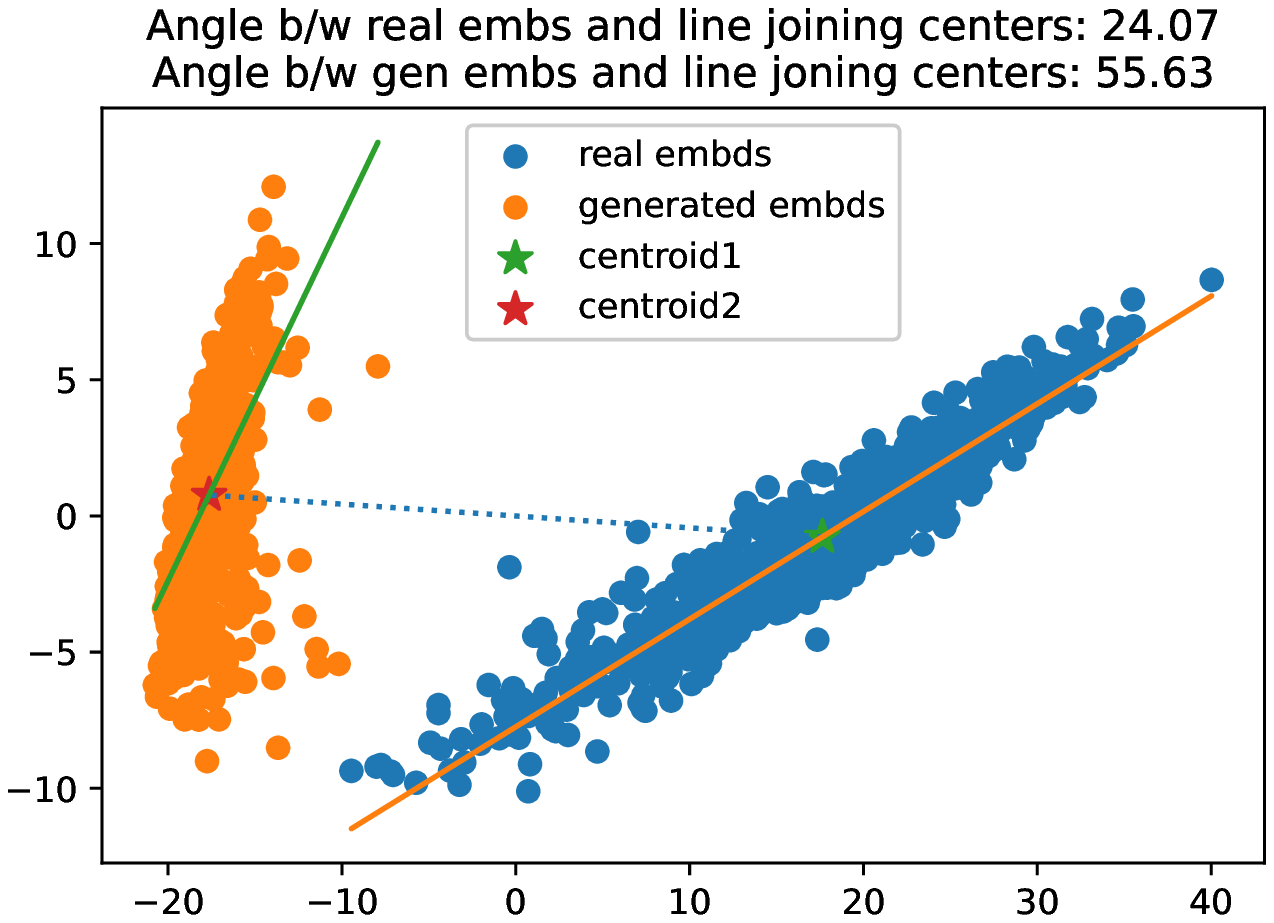}
     \end{subfigure} 
     \begin{subfigure}[b]{0.3\textwidth}
         \centering
         \includegraphics[width=\textwidth]{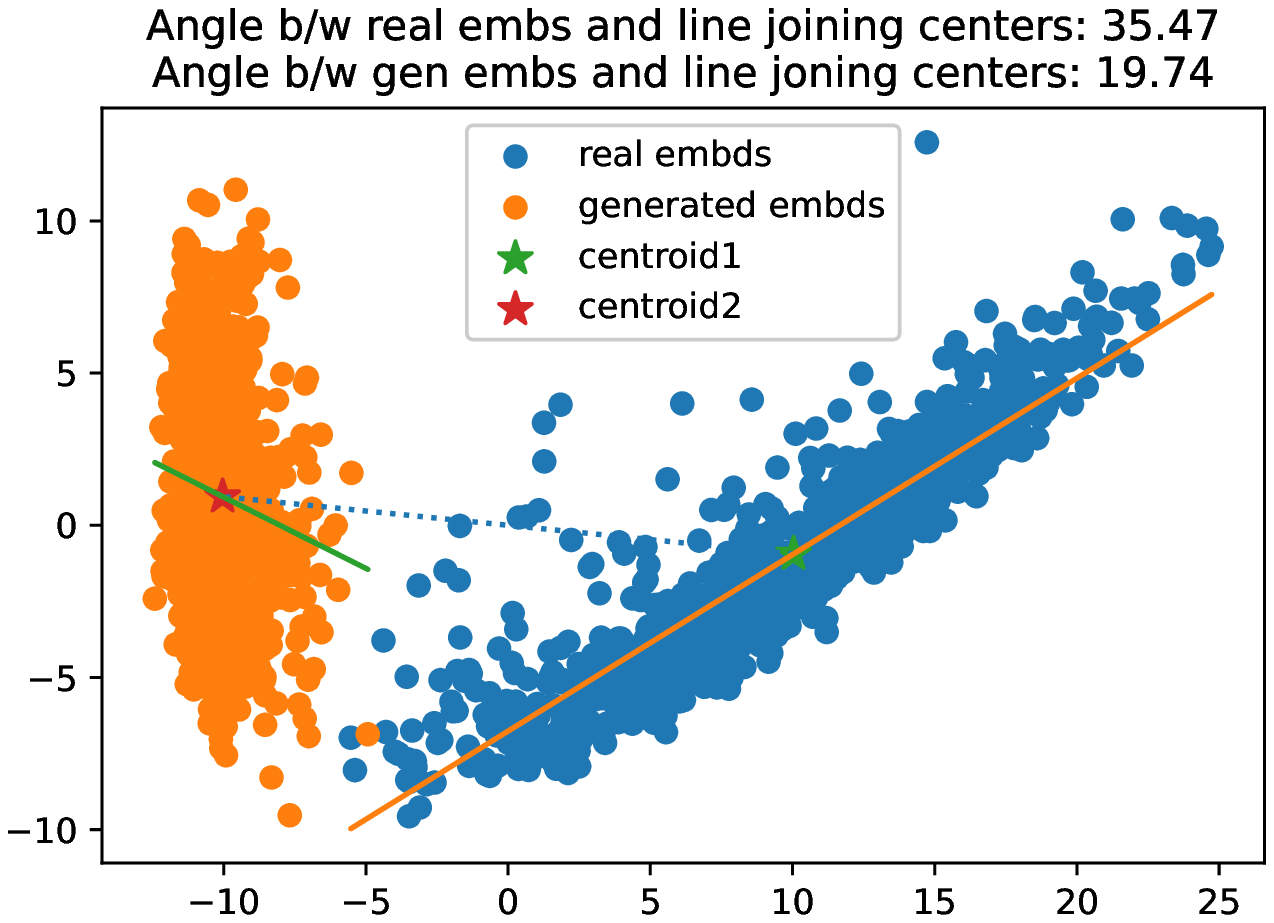}
     \end{subfigure}    
     \begin{subfigure}[b]{0.3\textwidth}
         \centering
         \includegraphics[width=\textwidth]{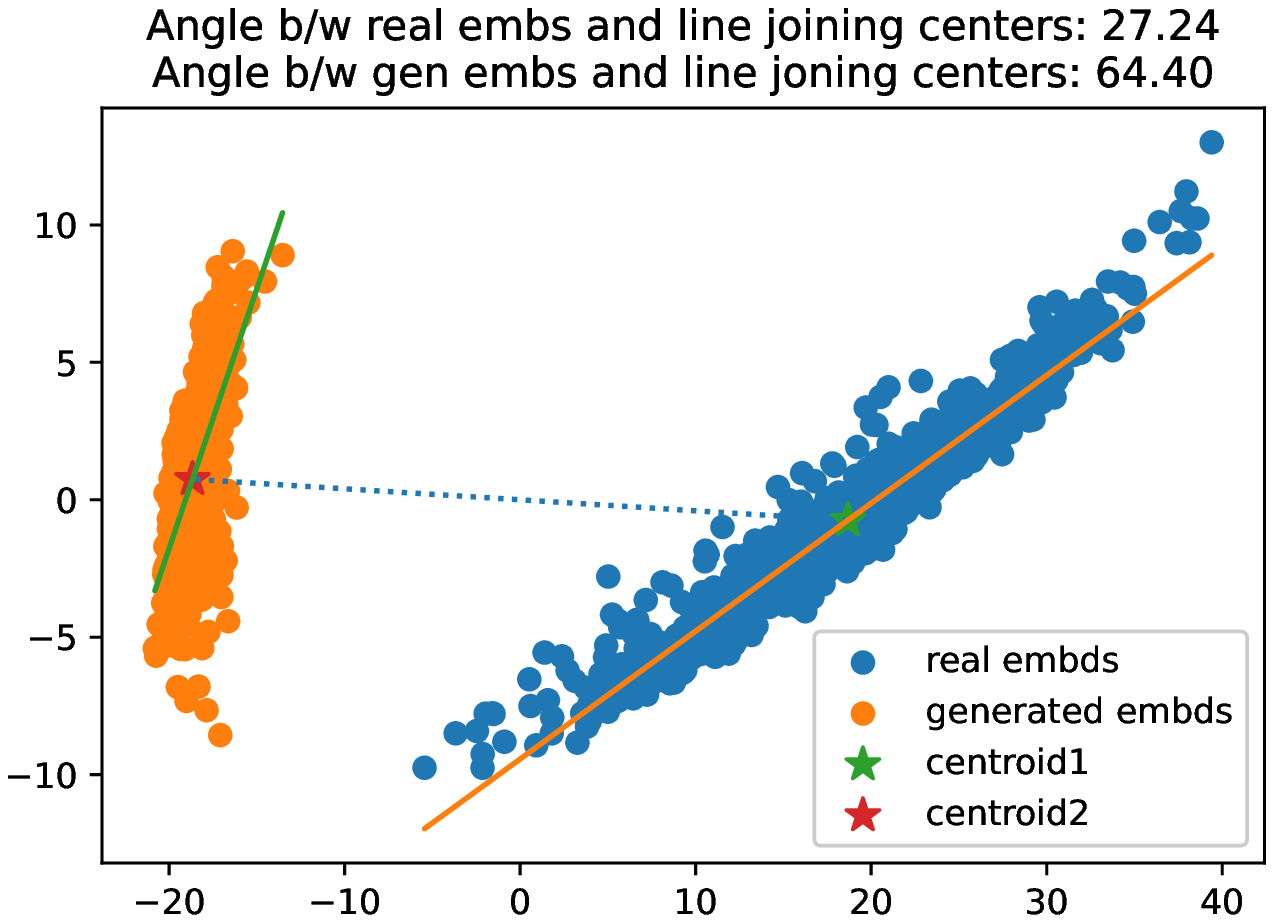}
     \end{subfigure} 
     \\
     \begin{subfigure}[b]{0.3\textwidth}
         \centering
         \includegraphics[width=\textwidth]{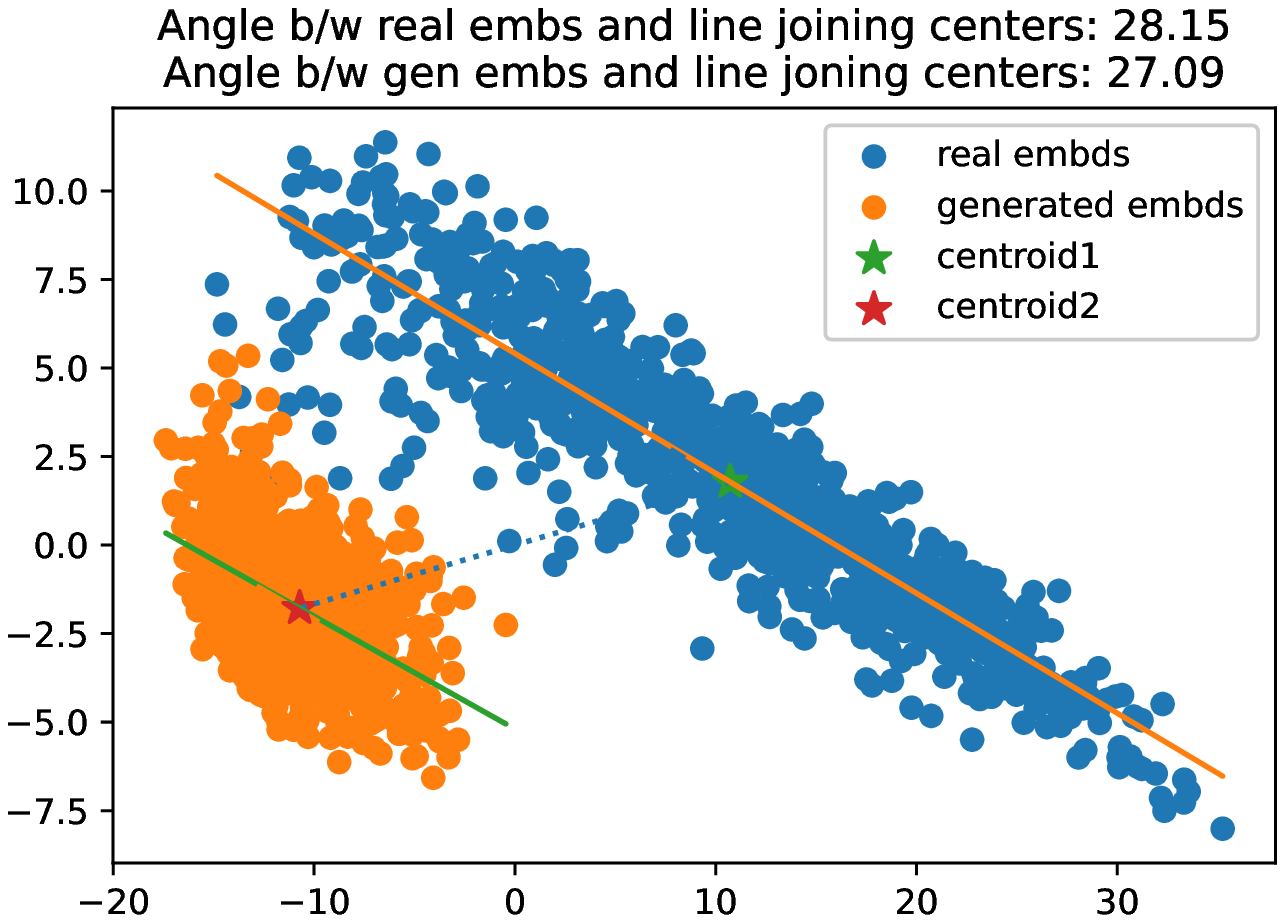}
     \end{subfigure}    
     \begin{subfigure}[b]{0.3\textwidth}
         \centering
         \includegraphics[width=\textwidth]{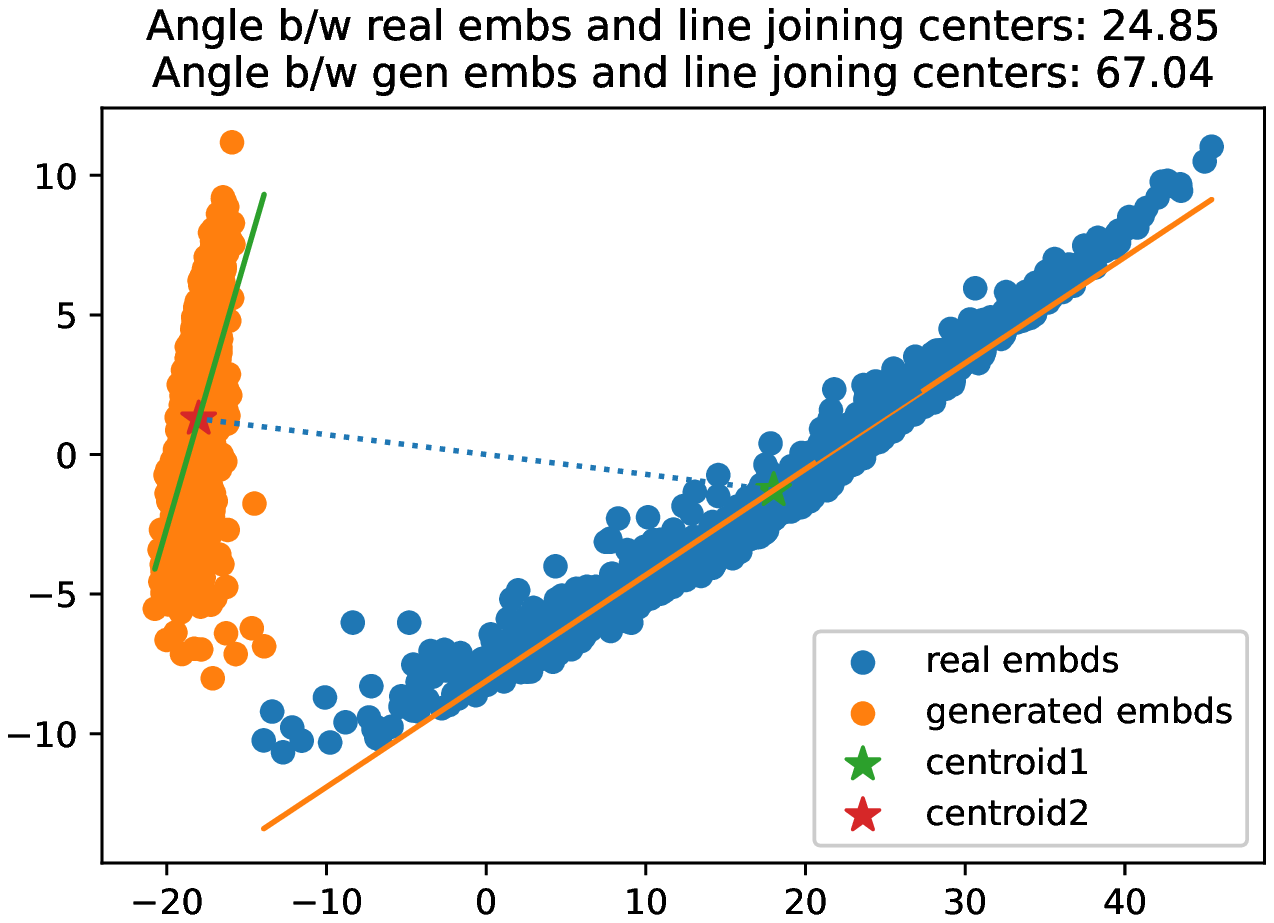}
     \end{subfigure}
     \begin{subfigure}[b]{0.3\textwidth}
         \centering
         \includegraphics[width=\textwidth]{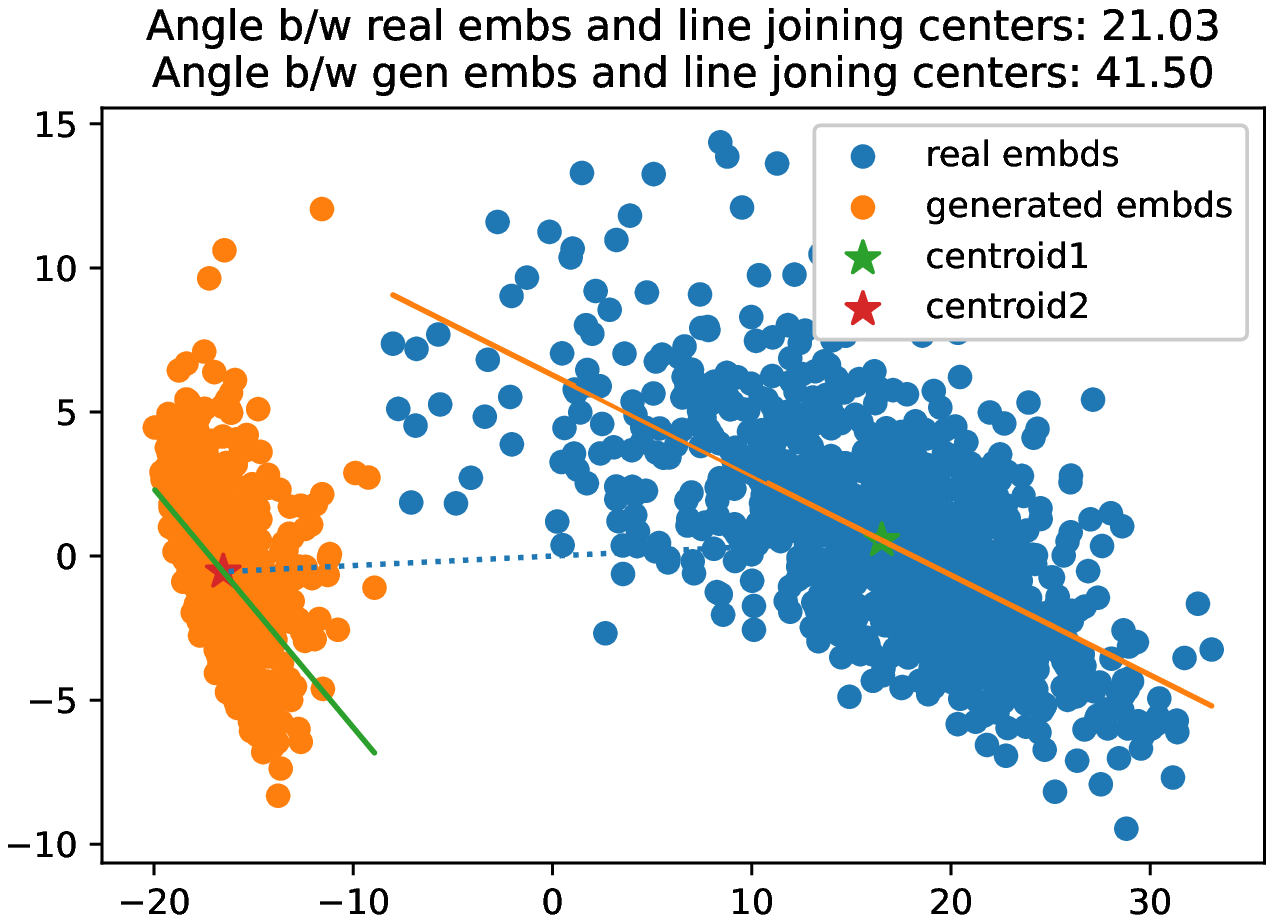}
     \end{subfigure}
        \caption{Illustration of 2D embedding (extracted from the penultimate layer of the discriminator) of nine different test images originally belonging to class `4' (blue) and generated images produced by the proposed method for target class `9' (orange) from the MNIST dataset. We present the line fittings along the principal axis of each embedding cluster depicting the direction of perturbation. The angles computed between the principal axes and the line joining the centroids indicate that adversarial samples follow well-defined trajectory from the original class towards target class in a majority of the cases.}
        \label{fig:lines_angles}
\end{figure*}

\begin{table}[h]
\centering
\caption{Comparison of the proposed method with FGSM and PGD in terms of classification accuracy on generated images (lower is better).}
\scalebox{0.75}{
\begin{tabular}{|lll||lll|}
\hline
\multicolumn{3}{|c||}{MNIST}                                                                                             & \multicolumn{3}{|c|}{CIFAR10}                                                                                             \\ \hline \hline
\begin{tabular}[c]{@{}l@{}}FGSM\\ $\epsilon$=0.1/0.2\end{tabular} & \begin{tabular}[c]{@{}l@{}}PGD\\ $\epsilon$=0.1/0.2\end{tabular} & Proposed & \begin{tabular}[c]{@{}l@{}}FGSM\\ $\epsilon$=0.01/0.02\end{tabular} & \begin{tabular}[c]{@{}l@{}}PGD\\ $\epsilon$=0.01/0.02\end{tabular} & Proposed \\ \hline
75.9/17.6                                                & 69.4/6.5                                                & 9.1     & 56.5/50.8                                                 & 5.7/0.1                                                 & 12.4   \\ \hline
\end{tabular}}
\label{PGD}
\end{table}

\subsection{Analysis}
\label{sec:analysis}

We examine how the class activations change after addition of adversarial perturbations using the proposed method. Although the original and perturbed images look visually alike, the class activation maps reveal complementary salient regions as shown in \autoref{fig:gradcam} on four different test images belonging to class label `0' on the CIFAR10 dataset. It is evident that the activations are sufficiently well-separated between the original images and the outputs generated by the proposed method.

We further visualize the embedding (2-D projection using t-SNE) of original and generated images,  extracted from the penultimate layer of $\mathcal{D}$. We select $\mathcal{D}$ over $\mathcal{G}$ as the embedding extractor because the images reconstructed from embedding produced by $\mathcal{G}'s$ encoder are already classified as the target class in a majority of cases. We present our analysis on nine different test images from the MNIST dataset originally belonging to class `4' and adversarially perturbed to target class `9' using the proposed method. We deliberately select these two classes as they are inherently difficult to distinguish compared to the rest of the digits. We observe in \autoref{fig:lines_angles} that the embedding of original and adversarial images form compact and disjoint clusters, indicating the proposed method successfully launched the attack in the latent space. Moreover, the embedding of generated images are aligned in a specific orientation with respect to the cluster of embedding from original images in a majority of cases, revealing a visually intuitive pattern in the trajectory of adversarial perturbations.

\section{Conclusion}
\label{sec:sum}
In this work, we design a GAN-based framework that induces adversarial perturbation in the latent space in contrast to existing methods that inject noise in the pixel space. The proposed method does not need an attack margin governing gradient-based attacks. Further, perturbations in the latent space can be geometrically interpreted using convex hulls. The proposed method achieves a reasonable performance in untargeted (up to 91\% attack success rate) and targeted attack (up to 93\% attack success rate) scenarios while maintaining sufficiently high degree of perceptual similarity between the original and adversarially perturbed images.

\clearpage
\bibliographystyle{unsrt}
\bibliography{egbib}
\end{document}